\documentclass[preprint,journal]{vgtc}            

\onlineid{1083}

\vgtccategory{Research}

\title{VisGuard: Securing Visualization Dissemination through \\ Tamper-Resistant Data Retrieval}

\author{%
  \authororcid{Huayuan Ye}{0009-0008-8208-2017},
  \authororcid{Juntong Chen}{0000-0001-9343-4032},
  \authororcid{Shenzhuo Zhang}{0009-0002-0638-6269},
  \authororcid{Yipeng Zhang}{0009-0003-5169-1008},
  \authororcid{Changbo Wang}{0000-0001-8940-6418},
  \authororcid{Chenhui Li}{0000-0001-9835-2650}
}

\authorfooter{
  \item
  	Huayuan Ye, Juntong Chen, Shenzhuo Zhang, Changbo Wang and Chenhui Li are with School of Computer Science and Technology, East China Normal University. Yipneg Zhang is with Institute of Education, Tsinghua University. Chenhui Li is the corresponding author. E-mail: chli@cs.ecnu.edu.cn.
}

\abstract{%
  The dissemination of visualizations is primarily in the form of raster images, which often results in the loss of critical information such as source code, interactive features, and metadata. While previous methods have proposed embedding metadata into images to facilitate Visualization Image Data Retrieval (VIDR), most existing methods lack practicability since they are fragile to common image tampering during online distribution such as cropping and editing. \sidecomment{SR.2 \\ R1.1}\revision{To address this issue, we propose VisGuard, a tamper-resistant VIDR framework that reliably embeds metadata link into visualization images. The embedded data link remains recoverable even after substantial tampering upon images.} We propose several techniques to enhance robustness, including repetitive data tiling, invertible information broadcasting, and an anchor-based scheme for crop localization. VisGuard enables various applications, including interactive chart reconstruction, tampering detection, and copyright protection. We conduct comprehensive experiments on VisGuard's superior performance in data retrieval accuracy, embedding capacity, and security against tampering and steganalysis, demonstrating VisGuard's competence in facilitating and safeguarding visualization dissemination and information conveyance.
}


\teaser{
  \centering
  \includegraphics[width=0.95\linewidth]{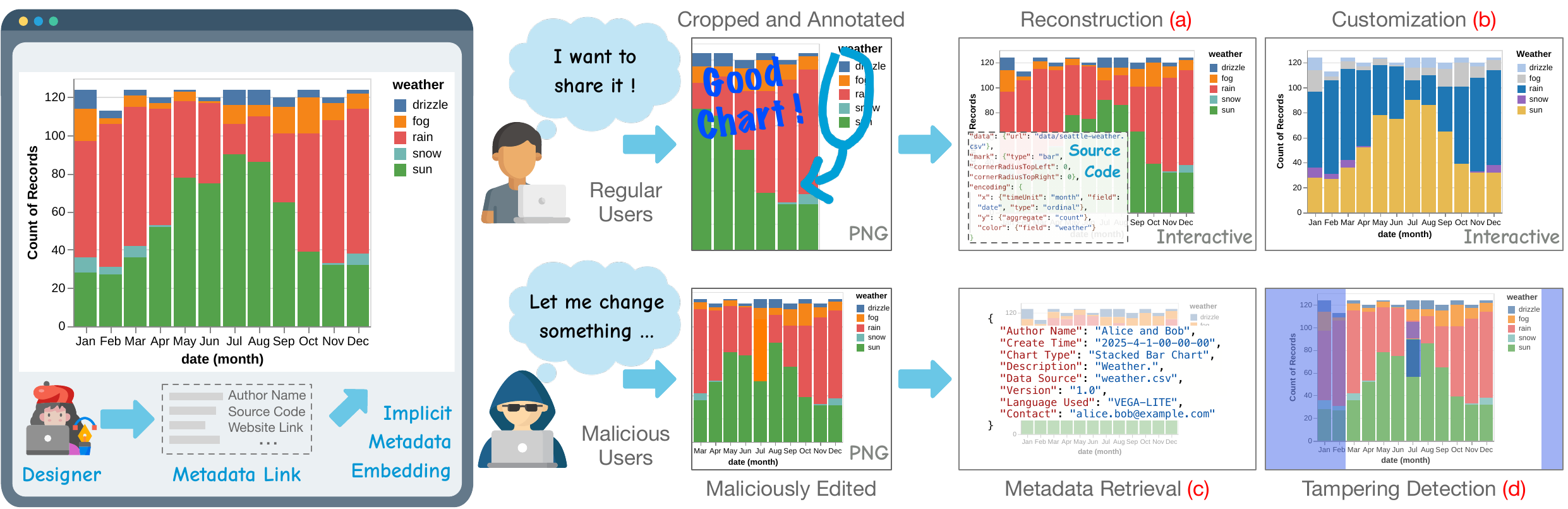}\vspace{-6pt}
  \caption{
  	  VisGuard is able to implicitly embed data into visualization images through a data link. The embedded data can be accurately restored even if the image is intentionally or unintentionally tampered with. With the decoded data, users can perform abundant explorations. For example, they can use the source code to reconstruct the interactive visualization~(a) or further customize its visual encodings~(b). Also, the metadata of the chart can be retrieved~(c) for copyright protection and information provenance. More importantly, the tampered areas can be detected~(d) to ensure the image has not been maliciously altered during dissemination.
  }
  \label{fig: teaser}
}

\graphicspath{{figs/}{build_in_figs/}} 

\usepackage[left=0.8cm,top=1.7cm,bottom=1.6cm,right=0.8cm]{geometry}

\usepackage{tabu}                      
\usepackage{booktabs}                  
\usepackage{lipsum}                    
\usepackage{mwe}                       
\usepackage{amsmath}
\usepackage{algorithm}
\usepackage{algpseudocode}

\usepackage{mathptmx}                  
\usepackage{fancybox}
\usepackage{adjustbox}
\usepackage{marginnote}
\usepackage{fancybox}
\usepackage{multirow}
\usepackage{multicol}
\usepackage{array}

\newcommand{\best}[1]{\textcolor{red}{#1}}
\newcommand{\sbest}[1]{\textcolor{blue}{#1}}

\begin{document}

\global\marginparsep=3pt
\newcommand{\sidecomment}[1]{%
  \ifdefined\revise%
  \marginnote{%
    \textcolor{Magenta}{%
      \adjustbox{minipage=0.2\marginparwidth,fbox}{%
          \tiny#1%
      }%
    }%
  }%
  \fi%
}%

\newcommand{\revision}[1]{%
  \ifdefined\revise%
    {\hypersetup{allcolors=Magenta}\textcolor{Magenta}{#1}}%
  \else%
    #1%
  \fi%
}%

\newcommand{\revisioncaption}[1]{%
  \ifdefined\revise%
    \textcolor{Magenta}{#1}%
  \else%
    #1%
  \fi%
}%

\newcommand{\revisionbox}[1]{%
  \ifdefined\revise%
  \setlength{\fboxrule}{0.6pt}%
  \setlength{\fboxsep}{0.5pt}%
  \fcolorbox{Magenta}{white}{#1}%
  \else%
  #1%
  \fi%
}%



\firstsection{Introduction}
\maketitle

Currently, visualization charts are disseminated primarily in the form of raster images because of their convenience and widespread accessibility~\cite{ye2023invvis}. However, visualizations offer more value beyond the pixels. Interactive exploration enabled by underlying source code can provide more intuitive interpretations and facilitate efficient concept communication~\cite{tufte1983visual}. Related metadata such as contextual reference links, source website information, and copyright details of the original creators are also essential components of visualizations. Extracting such textual information from visualization images addresses the critical challenge we term \textbf{visualization image data retrieval} (VIDR), which has broad applications including copyright protection, provenance verification, and enhancing the reusability and reliability of visual data across various dissemination platforms.

Previous studies have explored various approaches to address the VIDR problem. One line of work~\cite{poco2017extracting, poco2017reverse, song2022graphdecoder,chen2019towards} focuses on pattern recognition techniques to extract visual encodings. Although these approaches can recover visible metadata such as chart type, visual elements, and color mappings, they often suffer from limited extraction accuracy and cannot access nonvisible metadata. Another line of work~\cite{zhang2020viscode, fu2020chartem, ye2023invvis, fu2022chartstamp} adopts a proactive method by leveraging image steganography to embed metadata directly into visualization images. \sidecomment{SR.2 \\ R1.1}\revision{Unlike digital watermarking, which is imposed explicitly, steganography embeds data while concealing the existence of the hidden information.} This allows for arbitrary data embedding without altering the visual appearance of visualization images and enables users to restore embedded data for subsequent use.

However, the security and robustness of VIDR remain largely underexplored. Tampering often occurs during the online dissemination of visualization images. Benign tampering, such as cropping or resizing, is often applied to optimize image display or reduce file size. Malicious tampering, which refers to intentional modifications aimed at distorting the visual message or modifying copyright information using tools such as Photoshop, also occurs frequently. Existing pattern recognition and steganography-based methods are unable to reliably recover embedded data once the image has been tampered with. This vulnerability significantly limits their practical applicability, leading to risks of misinformation and copyright conflicts.

\sidecomment{SR.3 \\ R3.1}\revision{Although early studies using traditional pixel-based modifications cannot achieve satisfying data embedding and retrieval quality, recent efforts in tamper-resistant image steganography from the deep learning community~\cite{zhang2024editguard, sander2024watermark} have shown potential}, yet they fall short with limited embedding capacity and insufficient robustness to complex manipulations. Moreover, these methods are optimized for natural images, which largely differ from visualization images characterized by large homogeneous regions and structured graphical elements, making them inapplicable for VIDR.

Motivated by these gaps, we propose \textbf{VisGuard}, a novel VIDR framework that reliably safeguards the entire lifecycle of visualization creation and dissemination through tamper-resistant deep \sidecomment{SR.2 \\ R1.1}steganography. \revision{As shown in \cref{fig: teaser}, at the time of publication, chart creators can embed metadata links that may contain author information, source code and webpages. The embedded link remains recoverable even after the image undergoes substantial tampering during online distribution.} VisGuard supports a wide range of applications, including the reconstruction of interactive charts from embedded source code, facilitating the editing or reuse of visualization charts, and the detection of image tampering by localizing modified regions. We propose a deep learning-based pipeline for highly robust data embedding and retrieval while preserving the visual appearance of the original chart. We introduce several submodules to address various technical challenges related to tamper resistance to increase robustness. We further outline a new scheme to handle image cropping, which not only ensures data recovery from cropped images with high accuracy but also precisely predicts the cropped region.

We conduct comprehensive experiments to evaluate VisGuard. Our results show that VisGuard outperforms existing methods in terms of visual and perceptual quality, data decoding accuracy and security against steganalysis detection. \sidecomment{SR.4 \\ R3.3}\revision{Our method has high resistance to local tampering, image cropping and potential robustness to image degradation.} In summary, VisGuard addresses a critical gap in current VIDR studies with significantly improved robustness and reliability. Our contributions are as follows:

\begin{itemize} 
    \setlength{\itemsep}{0pt}
    \setlength{\parsep}{0pt}
    \setlength{\parskip}{2pt}
    \item [(1)] We identify the potential and challenges of the tamper-resistant VIDR problem. We explore and implement a variety of application scenarios for this problem to demonstrate its importance.
    \item [(2)] We propose a deep steganography-based framework that can achieve reliable and robust VIDR that safeguards visualization distributions with significantly improved data embedding and decoding capabilities.
    \item [(3)] We conduct a series of evaluations to demonstrate the superiority of our method from various aspects, including data embedding and retrieval quality, security, and capacity.
\end{itemize}
\section{Related Work}

\begin{figure*}[t]
    \centering
    \includegraphics[width=0.98\linewidth]{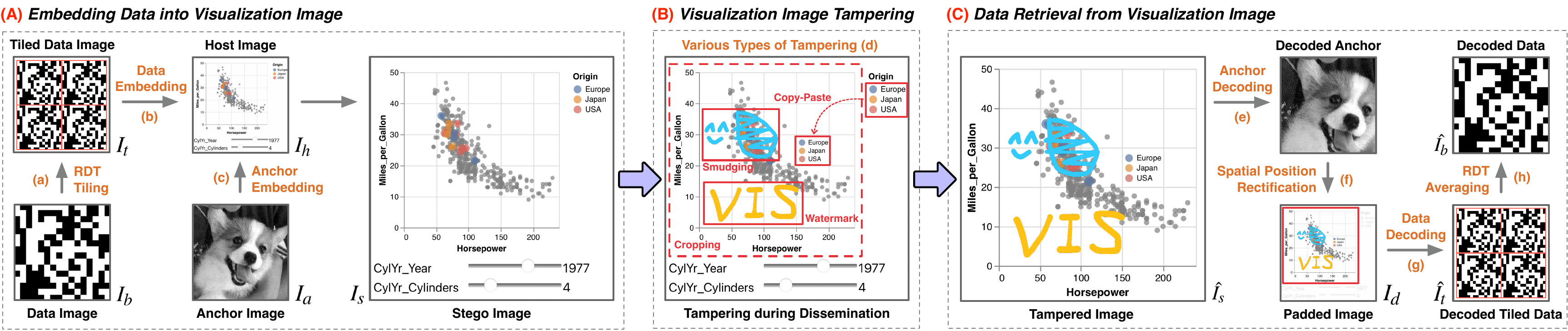}\vspace{-5pt}
    \caption{
        Given the data to be embedded (in the form of a binary image), it first undergoes repetitive data tiling~(RDT)~(a) to enhance information redundancy. Then, the tiled data~(b) and the anchor image~(c) used for cropping positioning are concealed in a host image through a steganography-based embedding network, resulting in a stego image. After that, the stego image can be disseminated and distributed on the internet and suffers various kinds of image tampering~(d). When receiving a potentially tampered image, users can subsequently decode the anchor image~(e) for cropping localization and recover its original relative position~(f), on which basis the tiled data can be decoded~(g) and the decoded data can be derived~(h).
   }
  \label{fig: pipeline}\vspace{-15pt}
\end{figure*}

\subsection{Information Steganography}

Information steganography aims to implicitly hide secret data in a carrier with unnoticeable changes. The carrier can include various kinds of data, such as text, images and videos. Delina et al.~\cite{delina2008information} proposed a text steganography scheme according to user-guided options. FontCode~\cite{xiao2018fontcode} hides data during font rendering with data space mapping. Yang et al.~\cite{yang20163d} proposed embedding information into 3D models with histogram adjustment. Delforouzi et al.~\cite{delforouzi2008adaptive} utilized the wavelet transform for audio steganography. \sidecomment{R2.3}\revision{Some stuidies~\cite{mao2024covert, mou2023large} implemented video steganography by hiding data in each frame.}

Because images serve as critical media for information transmission, numerous studies have focused on hiding information in images. Traditional methods generally embed information by modifying the spatial or transform domain of the host image~\cite{baluja2017hiding}. Spatial-domain methods mainly include least-significant-bit~(LSB) replacement~\cite{mielikainen2006lsb}, palette reordering~\cite{imaizumi2014multibit, niimi2002high} and bit plane complexity segmentation~(BPCS)-based schemes~\cite{kawaguchi1999principles, nguyen2006multi}. However, these types of methods can be easily detected by steganalysis techniques~\cite{fridrich2001detecting, yu2004reliable}. Some studies have proposed the use of high-level image features~\cite{pevny2010using} and distortion constraints~\cite{li2014new} to improve the method's security and undetectability. Transform-domain steganography hides data with different domain transformations, such as discrete cosine transform~(DCT)~\cite{almohammad2008high} and discrete wavelet transform~(DWT)~\cite{swanson1997multiresolution, zhu1999multiresolution}. However, the stego images generated by traditional schemes still have perceivable artifacts.
Recently, deep steganography, that which uses neural networks to implement information hiding, has achieved impressive performance. HiDDeN~\cite{zhu2018hidden} embeds binary messages into images through an autoencoder~(AE). Baluja et al.~\cite{baluja2017hiding} first hid a color image in another image by training an end-to-end network. Some methods~\cite{fu2022chartstamp, qin2020coverless, shi2017ssgan, tang2019cnn, tang2017automatic, zhang2019steganogan, tancik2020stegastamp} utilize adversarial training~\cite{goodfellow2014generative} to defend against steganalysis detection. More recently, the normalizing flow-based~\cite{dinh2014nice, dinh2016realnvp, kingma2018glow, ye2024pprsteg} model has shown promising performance in steganography tasks. This type of method can hide single~\cite{cheng2021iicnet, jing2021hinet, lu2021large} or multiple~\cite{guan2022deepmih, wu2021embedding} images in a host image. There are also several methods that focus on enhancing the robustness of steganography method against image distortions and manipulations, such as random noise~\cite{xu2022robust}, printing~\cite{fu2022chartstamp, tancik2020stegastamp, ye2024pprsteg, shadmand2024stampone} and inpainting~\cite{sander2024watermark}. 

Although existing methods can achieve high steganography quality, to the best of our knowledge, they either cannot resist image tampering or can only hide limited information. In this paper, we propose a novel framework that has higher tampering resistance and larger embedding capacity.

\subsection{Image Tampering Prevention}
Research on image tampering prevention has focused mainly on passive or proactive schemes~\cite{zhang2024omniguard}. The former aims to detect tampering by identifying anomalous regions such as artifacts, noise and resolution inconsistencies. MVSS-Net~\cite{dong2022mvss} uses multiview feature learning to detect image manipulations. HiFi-Net~\cite{guo2023hierarchical} uses a multilevel classifier to localize fine-grained tampering. Ma et al.~\cite{ma2023iml} introduced the transformer architecture~\cite{vaswani2017attention} to improve the detection accuracy. Yu et al.~\cite{yu2024diffforensics} leveraged diffusion prior~\cite{rombach2022high} for image forgery analysis. However, for intentional tampering upon visualization images that is almost invisible (e.g., copy-paste scatter points or number modification), existing methods cannot achieve a satisfactory performance.

Conversely, the proactive scheme embeds specific information into images with steganography methods, enabling the detection of tampering or the recovery of data via a decoder, even after the images have been altered. MaLP~\cite{asnani2023malp} embeds a learned template into images for tampering localization. StegaPos~\cite{egri2021stegapos} uses position field for cropping positioning. DRAW~\cite{hu2023draw} considers camera-shot images and restricts manipulation at the source. EditGuard~\cite{zhang2024editguard} and OmniGuard~\cite{zhang2024omniguard} can embed binary code into images with robustness against malicious watermarking and AIGC modifications. WAM~\cite{sander2024watermark} transforms the steganography task to segmentation task and achieves strong tampering resistance. Most of the abovementioned methods are designed for natural images. For visualization images that generally contain many homogeneous areas, imperceptible data embedding becomes much more difficult. In this paper, we focus on the tamper-resistant data embedding and retrieval for this type of image.

\subsection{Visualization Images Data Retrieval}
\label{sec: rel_vidr}

Visualization image data retrieval involves obtaining meta-information (e.g., copyright, source code, hyperlink) or other kinds of underlying data (e.g., chart type, color mapping) from a bitmap image~\cite{zhang2020viscode}. This technique can unlock the information confined by the static image and facilitate information delivery. 

Research on VIDR mainly has focused on two main types of technical approaches. The first type of method directly extracts information from visualization images using pattern recognition techniques. Savva et al.~\cite{savva2011revision} and Poco et al.~\cite{poco2017reverse, poco2017extracting} proposed the use of  a machine learning-based pipeline to extract underlying data and visual encoding specifications from raster images. Some methods~\cite{flower2016validity, mendez2016ivolver} introduce human interactions to improve the extraction accuracy. Some studies have focused on certain types of visualizations, such as simple 2D plots~\cite{al2017machine}, timelines~\cite{chen2019towards} and graph visualization~\cite{song2022vividgraph, song2022graphdecoder}. Generally, this type of method cannot achieve enough extraction precision due to the diverse and sophisticated visualization design space. Another type of approach uses image steganography to embed metadata into chart images and restore their information via a decoding procedure. Chartem~\cite{fu2020chartem} hides binary code in visualization images by modifying the background pixel values. VisCode~\cite{zhang2020viscode} uses an end-to-end network for metadata embedding and leverages a pretrained visual saliency network to improve the perceptual quality. Hota et al.~\cite{hota2019embedding} proposed facilitating the reproduction of scientific visualizations by embedding metadata into PDF files. ChartStamp~\cite{fu2022chartstamp} enhances the embedding robustness against real-world image distortions. InvVis~\cite{ye2023invvis} converts chart data to data images to improve the embedding capacity. Although this type of approach can avoid the recognition accuracy issue by slightly sacrificing image quality, existing methods cannot survive image tampering, such as cropping, inpainting and smudging, which are common types of interference during visualization distribution and dissemination. In this paper, we propose a novel VIDR solution that can accurately retrieve the hidden data even if the image is largely modified, improving the reliability of visualization dissemination.

\section{Methods}

\subsection{Overview}
\label{sec: overview}

As discussed in \cref{sec: rel_vidr}, for existing steganography-based VIDR methods, image tampering during dissemination can damage the information embedded, making the data unrecoverable. In this paper, we propose VisGuard, a novel framework, to address the tamper-resistant VIDR problem. Specifically, we focus on two major types of image tampering:
\begin{itemize} [leftmargin=0.15in]
    \item \textbf{Local Tampering} This type of tampering modifies specific regions of an image by smudging, inpainting, etc.~(an example is shown in \cref{fig: pipeline} (B)). These alterations can corrupt or distort the pixel values that originally conceal the data. To handle this type of tampering, we propose repetitive data tiling for information redundancy enhancement and a steganography network with an invertible information broadcasting module, introduced in \cref{sec: data_embedding}.
    \item \textbf{Image Cropping} This type of tampering is very common during image transmission via screenshot and user editing. It is also very challenging for VIDR as it restructures the spatial position of the entire image and leads to extensive loss of information. To address this problem, we propose embedding an additional anchor image for spatial position rectification, introduced in \cref{sec: anchor_embedding}.
\end{itemize} 

\sidecomment{SR.2 \\ R1.1}\noindent\textbf{Problem Definition} \cref{fig: pipeline} shows the pipeline of VisGuard. \revision{Formally, the input of our framework contains a set of binary code representing the metadata link to embed, a host image $I_h$ as the data carrier and an anchor image $I_a$ for cropping resistance.} Initially, the binary code is converted to a data image $I_b$ containing a series of black-and-white modules, with the black modules representing 0 and the white modules representing 1. The data image then undergoes a repetitive data tiling operation for redundancy enhancement, which dirives a tiled image $I_t$. In the data embedding process, we use two embedding functions learned by neural networks $E_b(\cdot)$ and $E_a(\cdot)$ to hide $I_t$ and $I_a$ sequentially in $I_h$ and obtain a stego image $I_s$ with $I_s = E_a(E_b(I_h, I_t), I_a)$. After that, the stego images can suffer image tampering and become $\hat{I_s}$. In the decoding procedure, an anchor decoder $D_a(\cdot)$ first decodes the anchor image and then restores the spatial position of $\hat{I_s}$ on this basis, deriving a padded image $I_{d} = D_a(\hat{I_s})$. Finally, the recovered tiled image $\hat{I_t}$ is obtained with a data decoder $D_b(\cdot)$ by $\hat{I_t} = D_b(I_d)$ and the data image $\hat{I_b}$ can be restored. Our goal is to minimize the difference between $I_b$ and $\hat{I_b}$ to lower the data loss.

\subsection{Data Embedding and Retrieval}
\label{sec: data_embedding}

\subsubsection{Repetitive Data Tiling}

Because local tampering will corrupt the pixel values in the host image, the information hidden in the tampered areas may be incorrectly decoded. To address this problem, we propose repetitive data tiling~(RDT), which arranges the data image as a unit repeatedly to form a grid to enhance information redundancy. Assume that the original secret data image $I_b$ contains $h \times w$ black or white modules representing 0 and 1, \sidecomment{R1.5}\revision{for RDT with a repetition of $(c_h, c_w)$, the tiled result will contain $(c_h \times h, \, c_w \times w)$ modules.} In this case, a single module in $I_b$ will appear \revision{$c_h \times c_w$} times in the tiled data image. To make the subsequent descriptions more intuitive, \sidecomment{R1.5}\revision{we define an RDT process as $RDT(h, w, c_h, c_w)$}. An example is shown in \cref{fig: pipeline}~(a), where the original data image contains $18 \times 18$ modules and this RDT process can be represented as $RDT(18, 18, 2, 2)$. \sidecomment{SR.2 \\ R1.1}\revision{This is also the default RDT parameter setting adopted in this paper, with an embedding capacity of 324 bits.} The tiled data image is resized to the same size as the host image and then fed to the subsequent module.

In the decoding process, given the recovered tiled image, we compute the mean value of all \sidecomment{R1.5}\revision{$c_h \times c_w$} appearances for each module. The final decoding result is obtained by rounding the average to 0 or 1. This inverse procedure is called RDT averaging in this paper.

\subsubsection{Invertible Information Broadcasting}

\begin{figure}[]
    \centering
    \includegraphics[width=1.0\linewidth]{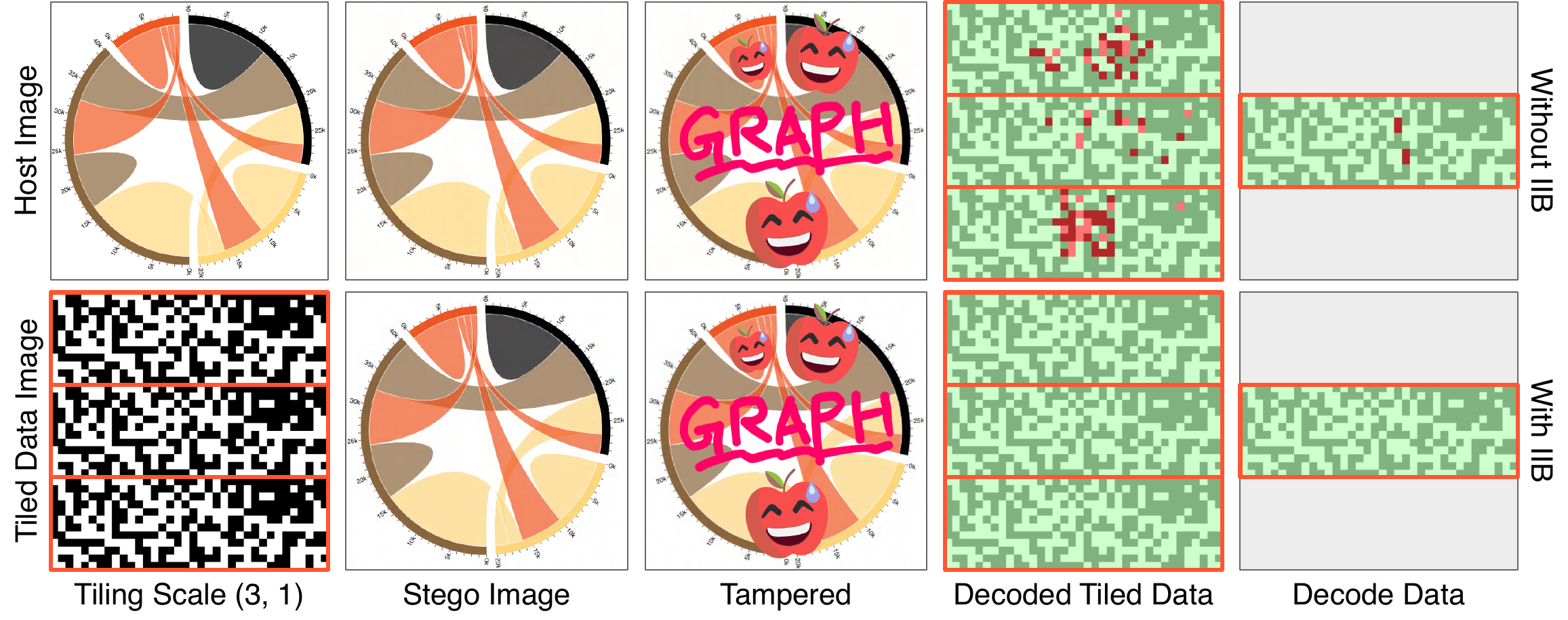}\vspace{-5pt}
    \caption{
        The data encoding and decoding results with (first row) and without IIB (second row). The green blocks represent correctly decoded data modules while the red ones indicate the opposite.
    }
    \label{fig: cmp_rdt_iib}\vspace{-15pt}
\end{figure}

\begin{figure*}[t]
    \centering
    \includegraphics[width=1\linewidth]{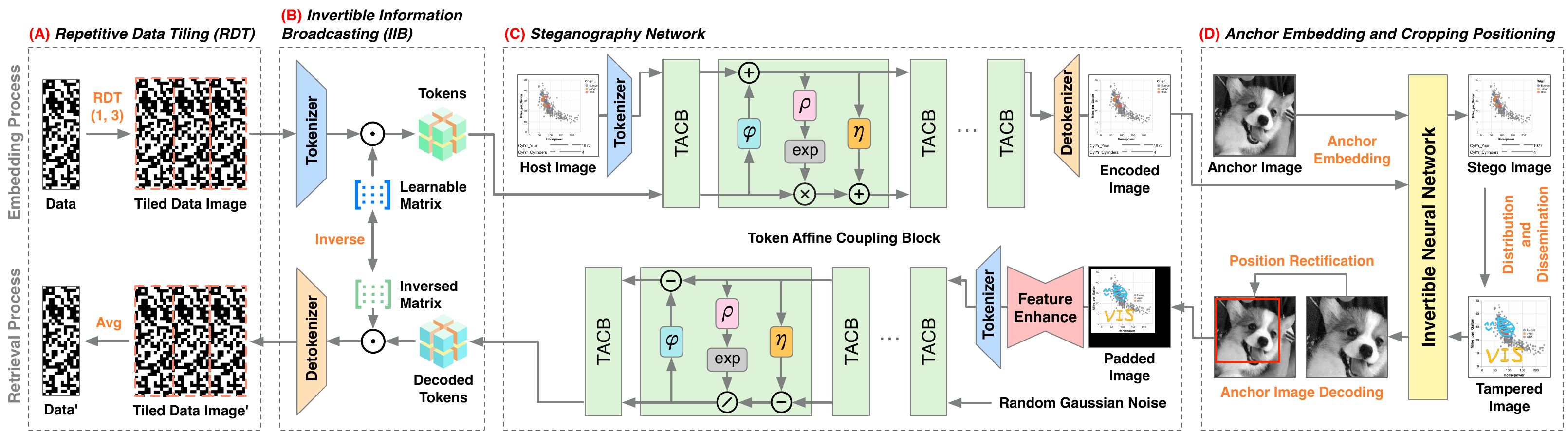}\vspace{-8pt}
    \caption{
        Model architecture of VisGuard. The upper part shows the data embedding process and the lower part shows the data retrieval process.
   }
  \label{fig: model}\vspace{-20pt}
\end{figure*}

Although RDT can effectively lower the possibility of data corruption, for large-area image interference~(including both local tampering and image cropping), the decoding accuracy will still significantly degrade. This degradation occurs because the bit data resides within a specific region, even though RDT makes it appear multiple times, the information still exists in a discrete form. Thus, once all locations where a data module appears are tampered with, its bit information will be completely lost. Based on this shortcoming, we propose the invertible information broadcasting~(IIB) module, whose design is shown in \cref{fig: model}~(B).

Given the tiled data image $I_t$ of size $(H, W)$, we first use a vision transformer~(ViT)~\cite{dosovitskiy2020image} as an image tokenizer to encode the image, deriving a tokenized tensor $T_t$ with shape $(N, D)$. Here $N$ indicates the token number and $D$ represents the latent dimension number. ViT can encode an image with attention scheme, during which the similarity score is calculated for tokenwise interaction. To further enhance the feature fusion among tokens, a learnable $N \times N$ matrix $M$ is applied to $T_t$ with matrix multiplication to obtain the transformed tensor $T_t^*$ by $T_t^* = M \cdot T_t$. In the data retrieval procedure, the inverse of $M$ is used to restore the tokens with $\hat{T_t} = M^{-1} \cdot \hat{T_t^*}$. The decoded tiled data image $\hat{I_t}$ is then derived by feeding $\hat{T_t}$ to a transformer decoder~(detokenizer).

As shown in \cref{fig: cmp_rdt_iib}, using IIB has almost no impact on the generated encoded image. For the same tampering, the decoded tiled data image without IIB has dense errors around the tampered regions, although the final error rate is effectively reduced after RDT averaging. In contrast, IIB can significantly improve decoding accuracy, as can be observed in this example, the result decoded with IIB achieves 100\% accuracy even in the tiled data image.

\subsubsection{Steganography Network}
Given the transformed data tokens $T_t^*$ output by the IIB module, we use a steganography network to embed them into the host image. Our network is based on the normalizing flow model~(NFM) proposed by Dinh et al.~\cite{dinh2014nice}. NFM, also known as the invertible neural network~(INN), allows rhe concealment and disclosure of data simultaneously in a single neural network by performing cascaded affine transformations with different calculation directions.

\cref{fig: model}~(C) demonstrates the architecture of our steganography network. In the data embedding process, we first tokenize the host image to $T_h$ with the same shape as $T_t^*$. Then, $T_h$ and $T_t^*$ are input to a series of token affine coupling blocks~(TACBs). Assuming that the inputs of the i\textsuperscript{th} TACB are $T_{h, \, i - 1}$ and $T_{t, \, i - 1}$, their corresponding outputs, $T_{h, \, i}$ and $T_{t, \, i}$, are obtained by applying the following affine transformations:
\vspace{-5pt}\begin{equation}\label{eq: tacb_forward}\vspace{-5pt}
    \begin{split}
        T_{h, \, i} &= T_{h, \, i - 1} + \phi(T_{t, \, i - 1}) \\
        T_{t, \, i} &= \eta(T_{h, \, i}) + T_{t, \, i - 1} \odot exp(\rho(T_{h, \, i})),
    \end{split}
\end{equation}
where $\odot$ indicates the Hadamard product, $exp(\cdot)$ represents the exponential function and $\phi(\cdot)$, $\eta(\cdot)$ and $\rho(\cdot)$ can be arbitrary neural networks as function learners. Previous methods~\cite{ye2023invvis, jing2021hinet, xu2022robust, lu2021large} have generally used neural networks~(CNNs), e.g., DenseNet~\cite{huang2017densely}. However, CNNs limit the feature representation within image channels and generate perceptible artifacts in the generated image. As a result, we choose to follow the design proposed by Ye et al.~\cite{ye2024pprsteg}, which leverages multihead self-attention block~\cite{ali2021xcit} that can enhance spatial information interaction to learn the affine functions. After $n$ TACBs, the encoded image can be obtained by detokenizing $T_{h, \, n}$.

In the decoding procedure, the steganography network takes the padded image $I_d$ with its cropping position rectified by anchor image decoding~(\cref{fig: model}~(D)) as input and outputs the restored tokens to the inverse IIB process. Owing to pixel corruption caused by tampering during visualization image dissemination, some features of data embedding can be weakened or even destroyed. This will lower the data recovery accuracy. To address this problem, we propose the use of a feature enhancement network~(FEN) to compensate for the feature loss before feeding $I_d$ to TACBs. Specifically, we use a UNet++~\cite{zhou2018unet++}, which adopts a nested network design to perform feature reconstruction, thus recovering and enhancing the spatial features of $I_d$. After that, the enhanced result is tokenized to $\hat{T_h}_{, \, n}$ and inversely goes through the TACBs with the reversed affine transformations:
\vspace{-6pt}\begin{equation}\label{eq: tacb_backward}\vspace{-6pt}
    \begin{split}
        \hat{T_{t}}_{, \, i - 1} &= (\hat{T_t}_{, \, i} - \eta(\hat{T_h}_{, \, i})) \odot exp(-\rho(\hat{T_h}_{, \, i})) \\
        \hat{T_h}_{, \, i - 1} &= \hat{T_h}_{, \, i} - \phi(\hat{T_t}_{, \, i - 1}).
    \end{split}
\end{equation}
In particular, $\hat{T_t}_{, \, n}$ is initialized as a random Gaussian noise, as $T_{t, \, n}$ can be assumed to obey a Gaussian distribution~\cite{jing2021hinet}. Defined by the essence of NFM~\cite{dinh2014nice}, \cref{eq: tacb_backward} is derived by simply reformulating \cref{eq: tacb_forward}. Thus, these two sets of transformations can be implemented by exactly one set of function learners~($\phi(\cdot)$, $\eta(\cdot)$ and $\rho(\cdot)$) with shared model parameters. This means that we can optimize TACBs end-to-end instead of training two independent networks~\cite{tancik2020stegastamp, zhang2020viscode} for data concealment and disclosure to achieve a better performance. Finally, the output of the first TACB, $\hat{T_t}_{, \, 0}$, is fed to IIB as $\hat{T_t^*}$ for the subsequent decoding process.

\subsection{Anchor Embedding for Cropping Resistance}
\label{sec: anchor_embedding}

Unlike local tampering, image cropping is difficult to defend not only because of large-scale pixel information loss but also because users sometimes cannot determine whether an image has been cropped. Take the tampered image shown in \cref{fig: pipeline} as an example, from the view of users, they can only see the explicit smudging areas while failing to note that the image is also cropped. Conversely, cropped images lose their relative position within the original image, making the network unable to recover the embedded data directly. Previous methods~\cite{sander2024watermark, egri2021stegapos} cannot handle these two problems simultaneously. In this paper, we propose a novel solution that allows the detection and localization of image cropping, as well as data retrieval from cropped images.

\begin{figure}[]
    \centering
    \includegraphics[width=1.0\linewidth]{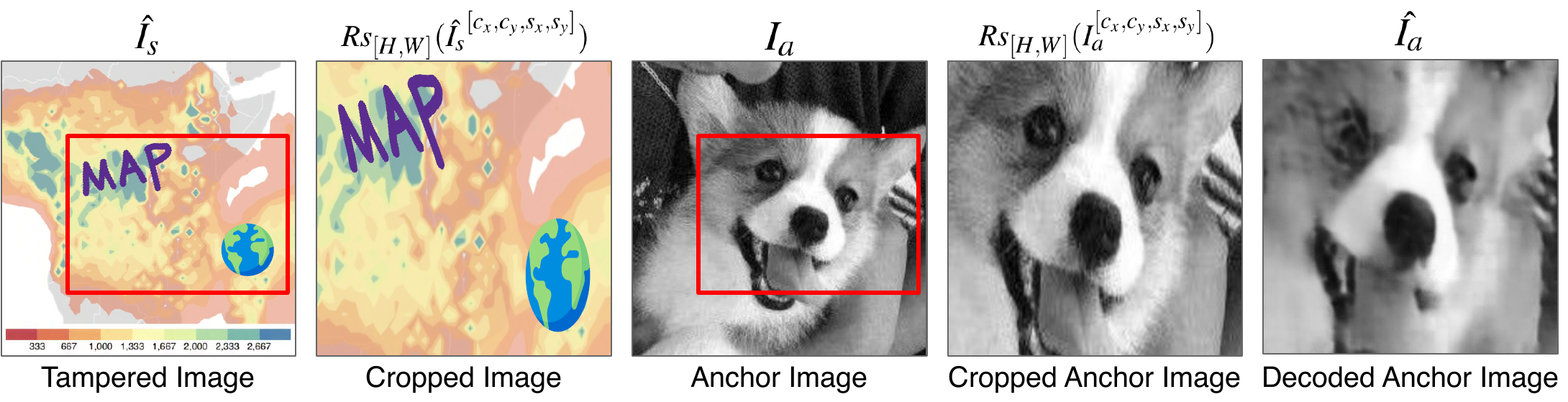}\vspace{-9pt}
    \caption{
        Demonstration of anchor decoding process. Symbols are based on the descriptions in \cref{sec: anchor_image_embedding}.
    }
    \label{fig: anchor_embedding}\vspace{-10pt}
\end{figure}

\subsubsection{Anchor Image Embedding}
\label{sec: anchor_image_embedding}

Given the image $I_e$ encoded with secret data, we aim to further embed a constant anchor image $I_a$ into it and derive a stego image as the final output. If the stego image is cropped during distribution and dissemination, its decoded anchor image will also be correspondingly changed. Hence, image cropping can be detected by comparing the difference between the original anchor image and the decoded image. This procedure is shown in \cref{fig: model}~(D).

To implement the embedding and decoding of anchor image, we utilize an off-the-shelf CNN-based NFM proposed by Lu et al.~\cite{lu2021large}. Here we do not adopt the transformer architecture used in our data steganography network as the anchor image information is supposed to maintain the consistency of its pixel positions in the stego image to respond correctly to image cropping. As a result, this embedding process should focus mainly on channelwise feature interaction, which is exactly what the CNN-based NFM excels at. Formally, the stego image $I_s$ is obtained by anchor embedding: $I_s = E_a(I_e, I_a)$ and the decoded anchor image $\hat{I_a}$ is derived through anchor decoding from the tampered image $\hat{I_s}$ (which is resized to the network input size) by $E^{-1}_a(\cdot)$, which is the inverse function of $E_a(\cdot)$ defined by the NFM with similar reformulation to \cref{eq: tacb_forward} and \cref{eq: tacb_backward}. During model training, assume that $\hat{I_s}$ has been cropped with $[c_x, c_y]$ as the center and $[s_x, s_y]$ as the scale, we apply the following constraint to guide the training of $E_a(\cdot)$:
\vspace{-5pt}\begin{equation}\label{eq: anchor_image_constraint}\vspace{-5pt}
    \mathcal{L}_{anchor} = \left\| \hat{I_a} - Rs_{[H, W]}(I_a^{[c_x, c_y, s_x, s_y]}) \right\|_1,
\end{equation}
where $Rs_{[H, W]}(\cdot)$ indicates resizing the image to the target size. As shown in \cref{fig: anchor_embedding}, by training the model with $\mathcal{L}_{anchor}$, the decoded anchor image will have the same cropping pattern as the tampered image. Although the decoded anchor image has some artifacts, mainly in regions that are locally tampered with, it is sufficient for cropping detection.

\begin{figure}[]
    \centering
    \includegraphics[width=1.0\linewidth]{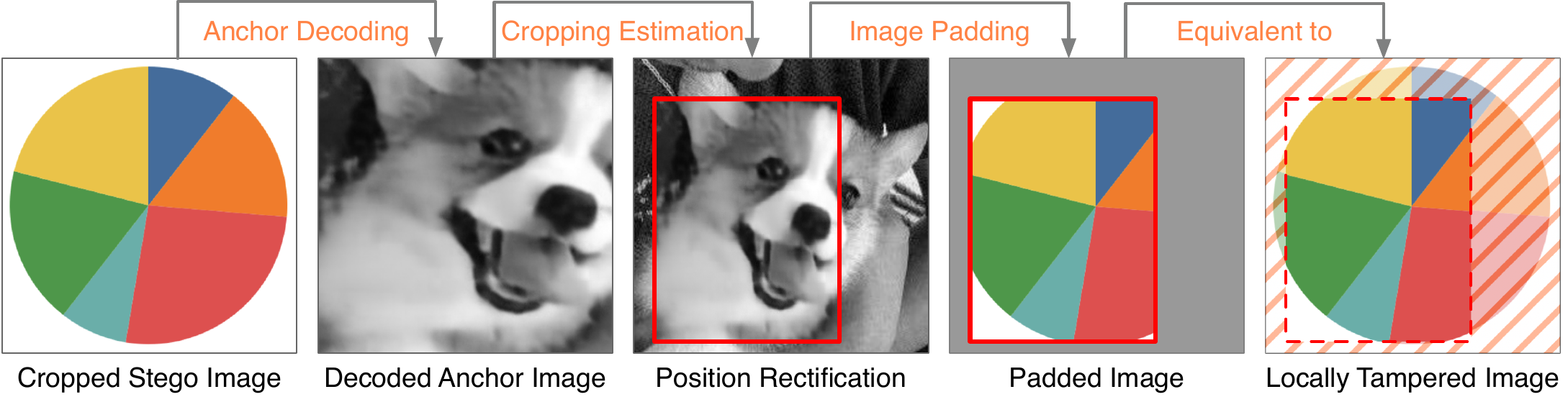}\vspace{-9pt}
    \caption{
        Demonstration of cropping estimation. Image cropping can be converted to local tampering via this scheme.
    }
    \label{fig: cropping_estimation}\vspace{-16pt}
\end{figure}

\subsubsection{Position Rectification via Cropping Estimation}
With the decoded anchor image $\hat{I_a}$, we can calculate its location in $I_a$ to estimate the position of the tampered image $\hat{I_s}$ within the stego image $I_s$, thereby obtaining the cropping parameters $[\hat{c_x}, \hat{c_y}, \hat{s_x}, \hat{s_y}]$. Then, we can rectify the relative position of $\hat{I_s}$ by padding the tampered image based on these parameters and thus converting the image manipulation from cropping to local tampering. As shown in \cref{fig: cropping_estimation}, the padded image is equivalent to the stego image whose inner edge is smudged.

To rectify the relative position of the tampered image $\hat{I_s}$, we model the cropping parameter estimation as an image template matching problem. Our goal is to obtain the best parameter set with:
\vspace{-5pt}\begin{equation}\label{eq: template_matching}\vspace{-5pt}
    \begin{split}
        \hat{c_x}, \hat{c_y}, \hat{s_x}, \hat{s_y} &= \arg \min_{\substack{c_x, c_y \\ s_x, s_y}} (\left\| \hat{I_a} - I_a' \right\|_1 + \lambda \cdot (1 - ssim(\hat{I_a}, I_a'))), \\[-5pt]
        I_a' &= Rs_{[H, W]}(I_a^{[c_x, c_y, s_x, s_y]}),
    \end{split}
\end{equation}
where $ssim(\cdot)$ indicates the structural similarity index~\cite{wang2004image} and $\lambda$ is a weight coefficient. \cref{eq: template_matching} aims to find a region in $I_a$ that best matches $\hat{I_a}$, thus determining the cropping parameters. In addition to pixel-level difference, here we also consider SSIM, as the decoded anchor image may exhibit blurring in certain areas due to local tampering~(an example is shown \cref{fig: anchor_embedding}). The incorporation of SSIM loss can improve the prediction accuracy by matching the overall structural information. In practice, \cref{eq: template_matching} can be optimized with gradient descent as this procedure is differentiable. With the predicted parameters, we can pad the tampered image and feed it to the data retrieval process.

\subsection{Training Strategy}

\subsubsection{Additive Stego Watermark}
\label{sec: additive_stego_watermark}
\sidecomment{SR.2 \\ R1.1}\revision{Because we incorporate the transformer architecture, the input and output image sizes of our network are required to be fixed~(384 $\times$ 384 in this paper) to support the attention mechanism.} However, in practical use, directly resizing the stego image can lead to insufficient resolution, especially when the original host image is large. As a result, we adopt the additive stego watermark strategy, which is also used by WAM~\cite{sander2024watermark}.

\begin{figure}[t]
    \centering
    \includegraphics[width=1.0\linewidth]{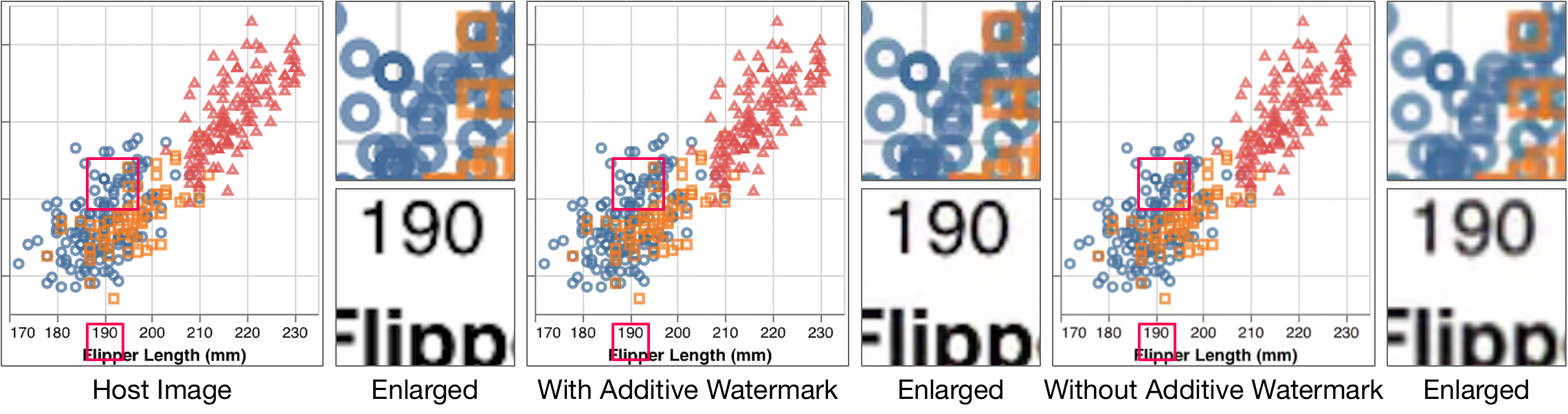}\vspace{-10pt}
    \caption{
        Comparison of the stego image generated using and not using additive steganography watermark. 
    }
    \label{fig: additive_watermark}\vspace{-20pt}
\end{figure}

\sidecomment{R1.6}\revision{Formally, given a host image in our training dataset (introduced in \cref{sec: experimental_settings})} whose size $I_h$ is $h_h \times w_h$, we first resize it to the network input size $h_n \times w_n$ to obtain the stego image $I_s$ with the same size. Then, we calculate the stego watermark, which is the difference between $I_s$ and $I_h$. Next, we resize this watermark to $h_h \times w_h$ and add it to the original host image to obtain the final stego image. By using an additive stego watermark, the stego image can have the same resolution as the host image, avoiding the issue of detail blurring that occurs when directly resizing from a low-resolution stego image to a high-resolution image. An example is shown in \cref{fig: additive_watermark}, where the stego image generated without using additive watermark has obvious blurring artifacts. This also makes our method easier to deploy in practical scenarios~(discussed in \cref{sec: app_source_end}).

\subsubsection{Tampering Simulation}
\label{sec: tampering simulation}
To make our method tamper-resistant, we simulate image tampering during the training process. Specifically, we apply different tampering operations to the generated stego images and then utilize these altered images for anchor decoding and data retrieval:

\begin{itemize}[leftmargin=0.15in]
    \item \textbf{No Operation} In this case, the stego image remains untampered.
    \item \textbf{Random Masking} To simulate malicious watermarking and smudging, we add a random mask to the stego images. In particular, we mask the image $k$ times, with each time overlaying a random region with a randomly sized texture cropped from another random image. In our implementation, $k \in [1, 4]$ and the proportion of unmasked areas is constrained to be no less than $\epsilon_m = 20\%$.
    \item \textbf{Random Cropping} As introduced in \cref{sec: anchor_embedding}, we can convert image cropping to local tampering. Hence, we perform random cropping only for the training of the anchor embedding and decoding network. Formally, we randomly crop the stego image and constrain the cropped image size to be no less than $\epsilon_c = 10\%$ of the original size.
    \item \textbf{Random Transition} In practice, cropping estimation may involve prediction errors. This can cause uncertain transitions within the padded image and thus lead to data decoding errors. As a result, we incorporate the transition simulation module proposed by StegaStamp~\cite{tancik2020stegastamp}. In this paper, we set the transition factor $\tau$ as 0.01.
\end{itemize}

During training, the above operations are applied with random combinations, e.g., a stego image can be cropped and masked simultaneously.

\subsubsection{Loss Function}
Our model is trained by jointly optimizing three parts: the IIB module, the data steganography network and the anchor embedding network. We guide the training process in three aspects: stego image quality, data decoding accuracy and anchor decoding accuracy. For stego image quality, we leverage the following hybrid loss function:
\vspace{-5pt}\begin{equation}\label{eq: loss_steg}\vspace{-5pt}
    \mathcal{L}_{steg} = \lambda_1 \cdot \left\| I_h - I_s \right\|_1 + \lambda_2 \cdot (1 - ssim(I_h, I_s)) + \lambda_3 \cdot lpips(I_h, I_s),
\end{equation}
in which $lpips(\cdot)$ represents the learned perceptual image patch similarity~(LPIPS)~\cite{zhang2018unreasonable}, which reflects the perceptual similarity between two images. For data decoding accuracy, we calculate the loss between the original data image and the restored data image~(after RDT averaging):
\vspace{-5pt}\begin{equation}\label{eq: loss_data}\vspace{-5pt}
    \mathcal{L}_{data} = \lambda_4 \cdot BCE(\hat{I_b}, I_b),
\end{equation}
where $BCE(\cdot)$ indicates the binary cross-entropy loss. For anchor decoding accuracy, as described in \cref{sec: anchor_embedding}, we use $\mathcal{L}_{anchor}$, which is introduced in \cref{eq: anchor_image_constraint}, to supervise the anchor decoding result.

Notably, the gradient of $\mathcal{L}_{data}$ is computed and propagated back if and only if random cropping is not performed on the stego images within a training batch. This is because the cropped stego image will not be directly fed into the steganography network for data decoding. Instead, we only input images with correct spatial relative positions, which can be either uncropped images or padded images. Thus, the overall loss function can be formulated as:
\vspace{-5pt}\begin{equation}\label{eq: loss_total}\vspace{-5pt}
    \mathcal{L}_{total} = \mathcal{L}_{steg} + \lambda_{crop} \cdot \mathcal{L}_{data} + \lambda_5 \cdot \mathcal{L}_{anchor},
\end{equation}
where $\lambda_{crop}$ is 1 if random cropping is not used and 0 otherwise. In addition, $\lambda_i$ in \cref{eq: loss_steg}, \cref{eq: loss_data} and \cref{eq: loss_total} are weight coefficients.

\section{Applications}

\subsection{Robust Invertible Visualization}

The concept of invertible visualization was first defined by InvVis~\cite{ye2023invvis}. It involves restoring or further modifying the visualization from an image. Given that visualizations are generally disseminated through the form of raster images (e.g., screenshots), this technique can enable users to reobtain interactive charts and help them understand the underlying data. Previous methods have explored many application scenarios. However, they are generally very susceptible to image tampering, which is almost unavoidable during the transmission of visualization images. As a result, improving the robustness of the invertible visualization framework is very meaningful for practical use.

\begin{figure}[h]
    \centering
    \includegraphics[width=1.0\linewidth]{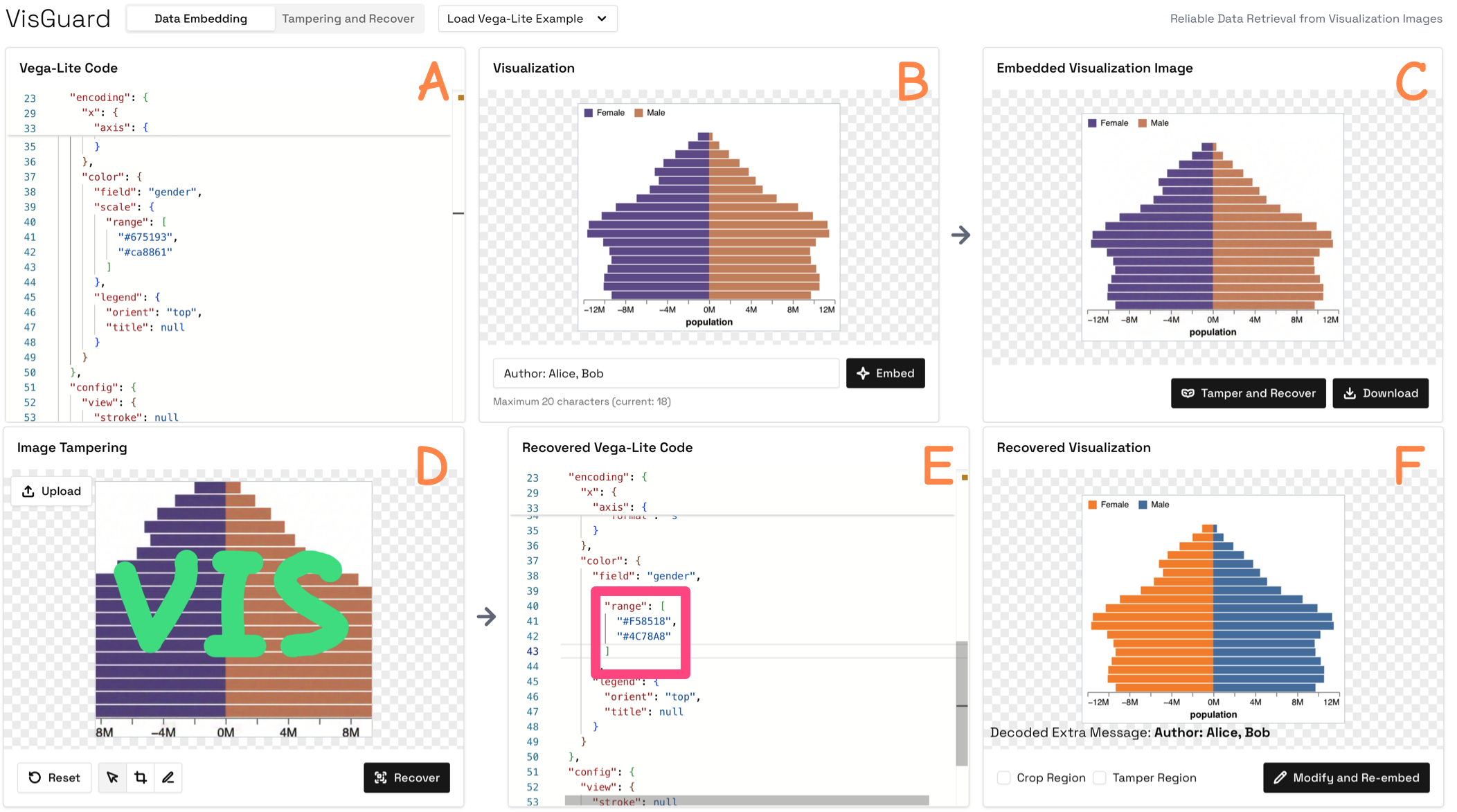}\vspace{-6pt}
    \caption{
        Application interface for invertible visualization. Uses can create or edit a visualization~(A) with code~(Vega-Lite~\cite{satyanarayan2016vega} in this case). The source code will be represented through a link and embedded into the visualization together with the additional information input by users~(B) and produce a stego image that can be distributed~(C). When a user receive a chart image, which can possibly be a tampered one~(D), they can retrieve the embedded data and use the source code~(E) to reconstruct or modify the visualization~(F).
    }
    \label{fig: app_inv_vis}\vspace{-17pt}
\end{figure}


VisGuard can solve the robust invertible visualization problem with its tamper-resistant property. We design our pipeline by placing greater emphasis on robustness than on embedding capacity, as in practical applications, large volumes of data~(e.g., json files or images) can be effectively represented through a single link. Once the link is successfully decoded, all the data behind it are available. As shown in \cref{fig: app_inv_vis}, we develop a website where users can either create visualizations and embed source code and other information, such as copyright, into it or decode the data and then use it for invertible visualization. In this case, we embed the source code link and author information into a chart image. Despite the image being heavily tampered with during dissemination, users are still able to decode the correct link using VisGuard. Through this link, they can access the source code of the visualization, reconstruct an interactive chart and further conduct personalized modifications. For example, in \cref{fig: app_inv_vis}~(E) and (F), we modify the decoded source code and change the colors of the chart. In our implementation, we use BCH code to perform error correction on binary code, thereby enhancing the accuracy of information restoration.

\subsection{Visualization Tampering Detection and Localization}
Visualization is an important and efficient way to convey quantitative data~\cite{tufte1983visual}. However, when visualizations are disseminated online in the form of images, their ability to convey knowledge can become a double-edged sword. This is because images can be easily altered through various image editing applications, even simple screenshot tools allow effortless image cropping and annotation. Some image alterations are unintentional, such as repositioning the legend or cropping a section to accentuate the focal point. Others, however, are malicious, such as imposing watermarks that disrupt visual encodings or even tampering with the data within the chart. Regardless of their intent, these modifications manipulate the original design created by the visualization creator. Certain changes can be particularly deceptive, leading users to assimilate incorrect information, which contradicts the fundamental principle of informativeness that visualizations are meant to embody. As a result, it is meaningful if users can confirm whether and where a received visualization image has been tampered with.

\begin{figure}[t]
    \centering
    \includegraphics[width=1.0\linewidth]{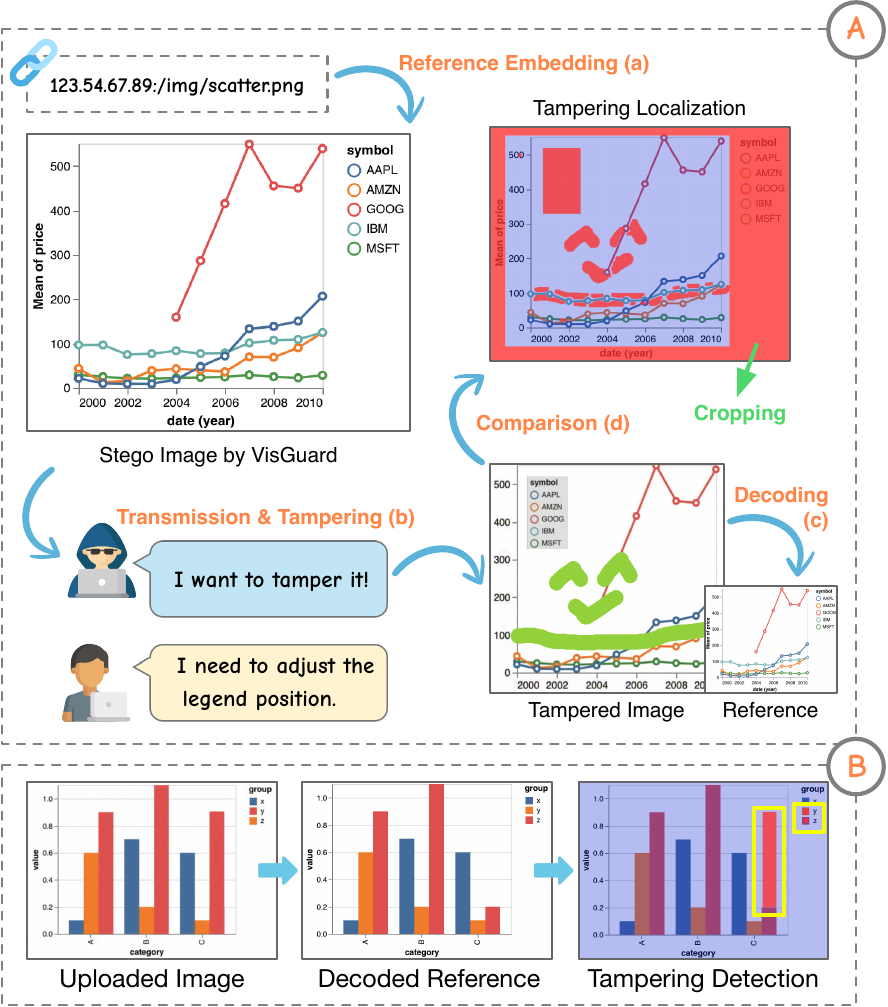}\vspace{-7pt}
    \caption{
        \textbf{(A)}: How VisGuard supports visualization tampering detection and localization. First, A stego image is generated by embedding a reference image~(a). Then, the stego image undergoes dissemination and tampering~(b). When receiving an image, users can decode the link to obtain the reference image~(c) and compare it with the received one~(d) to identify whether and where the image has been tampered. \textbf{(B)}: VisGuard can detect subtle tampering that is visually invisible. 
    }
    \label{fig: app_tampering}\vspace{-20pt}
\end{figure}

Because VisGuard can accurately decode the embedded data even if the stego image is tampered with, it can provide a novel solution to this problem. As shown in \cref{fig: app_tampering}~(A), we embed a link to the reference image~(the same as the host image) into the host image. Users can obtain the reference image by decoding the link. The differences between two images can then be calculated and users can localize the tampering. As shown in \cref{fig: app_tampering}~(B), in this case, the stego image is maliciously tampered with subtle modifications by adjusting the bar height and switching the legends. This type of tampering implicitly obscures the data lying behind the chart and can cause misleading. Using VisGuard, visually invisible manipulations can be easily detected to guarantee the reliability of the visualization dissemination.

\begin{figure}[b]\vspace{-18pt}
    \centering
    \includegraphics[width=1.0\linewidth]{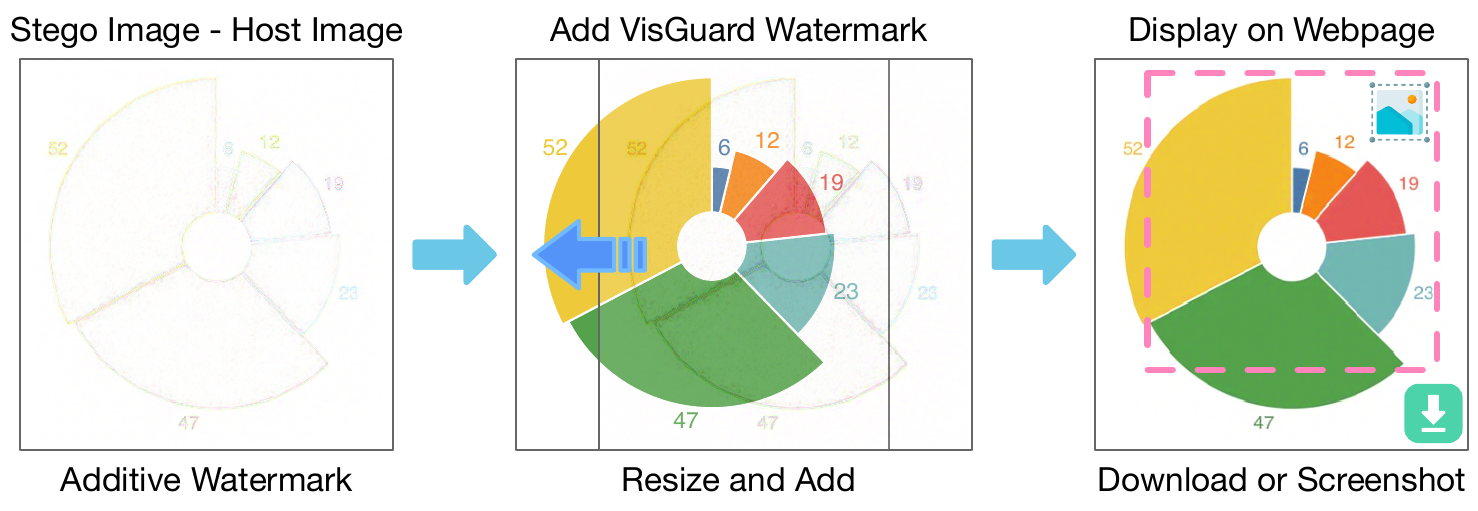}\vspace{-10pt}
    \caption{
        Our watermark-based pipeline to implement SEVDE. The watermark is enhanced by 3 times for a better illustration.
    }
    \label{fig: app_source_end}
\end{figure}

\begin{figure*}[]
    \centering
    \includegraphics[width=0.98\linewidth]{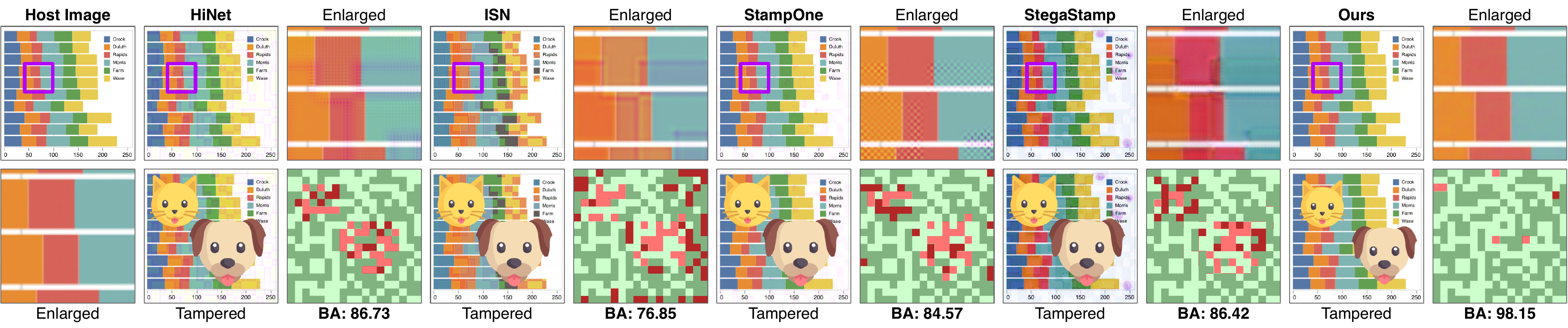}\vspace{-11pt}
    \caption{
        Data embedding quality compared with \textsl{cat.A} methods. The green and red blocks represent correctly and incorrectly decoded data modules, respectively.
   }
  \label{fig: evl_quality_cat_a}\vspace{-17pt}
\end{figure*}

\subsection{Source-End Visualization Data Embedding}
\label{sec: app_source_end}
Another critical issue of visualization data embedding is that ordinary users typically do not proactively embed data into chart images, as they are disseminators rather than creators of information and they lack sufficient motivation or need to perform such an operation. For them, embedding data does not yield direct benefits or value. In contrast, this need primarily stems from the authors or publishers of the charts, who aim to ensure the integrity and traceability of the charts by embedding data, or to provide an additional layer of information through invertible visualization, enabling users to delve deeper into and analyze the underlying data behind the charts. Therefore, only by embedding data into visualizations at the source can we ensure that the chart images disseminated across the internet are equipped with embedded data. We call this source-end visualization data embedding~(SEVDE). A core challenge of implementing SEVDE lies in ensuring that all chart images obtained from webpages should carry embedded data and meanwhile the data can be accurately retrieved. Previous methods only work when images are intact, i.e., direct download. However, in some cases. users may choose more convenient methods, such as screenshots, which allow them to freely select the captured area. In such scenarios, existing steganography-based VIDR methods fail to extract the embedded data, posing a major challenge in implementing SEVDE.

We design a novel pipeline to implement SEVDE with VisGuard. Specifically, we adopt a implicit watermark-based solution, which is shown in \cref{fig: app_source_end}. \sidecomment{R1.3}\revision{When a visualization designer finishes creating a chart, their customized information, together with the autogenerated identifiers (invisible to common users) such as authorship data and timestamps to ensure data integrity, can be converted to a corresponding additive stego watermark (introduced in Sec.~\ref{sec: additive_stego_watermark})}. After that, the stego watermark is resized~(to match the resolution) and added to the visualziation and displayed to users. Through this method, charts displayed on the webpage will carry VisGuard watermarks, ensuring that users who screenshot or download the chart obtain an implicitly watermarked result. Because VisGuard is tamper-resistant, even if the screenshot contains only a part of the visualization, the embedded data can still be decoded. This design eliminates the need for ordinary users to proactively embed data by making the embedding process finished at the source end and only requires visualization publishers to add stego watermarks to their charts. In practical use, this method is plug-and-play, as the stego watermark can be calculated efficiently as a postprocess of visualization creation. This means that it can support any visualization authoring tools, such as D3~\cite{bostock2011d3} and Vega-Lite~\cite{satyanarayan2016vega}. In summary, this is a win-win strategy: not only can visualization authors easily embed data to achieve functions such as copyright \sidecomment{R1.3}protection, \revision{but users can also easily verify whether the images they receive are derived from the original (by validating the decoded identifiers) and unaltered visualizations and use the decoded data for further explorations.}
\section{Evaluation}

\subsection{Experimental Settings}
\label{sec: experimental_settings}
\textbf{Datasets} Because VisGuard is designed for visualization images, natural image datasets are not suitable for our task. Instead, we compose a dataset that comprises a total of 21,782 images by incorporating two public datasets: InfoVIF~\cite{lu2021modeling} and MASSVIS~\cite{borkin2015beyond}. InfoVIF consists primarily of infographics, whereas MASSVIS contains mainly chart images. Our training and testing sets are derived by splitting this combined dataset into a 5:1 ratio. Compared with previous methods~\cite{ye2023invvis,zhang2020viscode}, we significantly expand the dataset size in this paper, as transformer-based models require a larger volume of training data to achieve better convergence and performance~\cite{kaplan2020scaling, dosovitskiy2020image}. The secret data used for embedding are randomly generated during training and the anchor image is predefined with an arbitrary choice.

\textbf{Model Implementation} Our full model is trained on 4 NVIDIA GeForce 3090 GPUs with an image size of $384 \times 384$ for network input. The patch size for all the attention calculations is $16 \times 16$ and the token dimension is 768. We use 4 TACBs for our full model. We use $RDT(18, 18, 2, 2)$ for data preprocessing, which means that the model has an embedding capacity of 324 bits. Our model is implemented with PyTorch~\cite{paszke2019pytorch} and optimized with the AdamW optimizer~\cite{loshchilov2017decoupled}. The initial learning rate is set as 0.0001 and it decays by $10\%$ after each epoch. \sidecomment{R1.4}\revision{The hyperparameters are set as $\lambda = 0.1$, $\lambda_1 = 1.0$, $\lambda_2 = 0.01$, $\lambda_3 = 0.1$, $\lambda_4 = 0.6$ and $\lambda_5 = 0.025$, based on an empirical configuration that achieves a relatively balanced model performance according to our experiments.} The model is trained for 30K iterations, which takes approximately 8 hours.

\textbf{Baselines} To perform a comprehensive evaluation, we select two categories of baselines. The first category~(\textsl{cat.A}) consists of vanilla image-in-image steganography methods: HiNet~\cite{jing2021hinet}, ISN~\cite{lu2021large}, StegaStamp~\cite{tancik2020stegastamp} and StampOne~\cite{shadmand2024stampone}. Although these methods are not specifically designed for tamper-resistant data embedding, they can generate high-quality stego images encoded with large volumes of data. The second category~(\textsl{cat.B}) includes tamper-resistant methods: WAM~\cite{sander2024watermark} and EditGuard~\cite{zhang2024editguard}. These two methods directly encode binary data into the host image without using data image as medium, achieving tamper resistance at the cost of embedding capacity. To ensure a fair comparison, we retrain the models of \textsl{cat.A} and \textsl{cat.B} methods on our dataset using the tampering simulation introduced in \cref{sec: tampering simulation}. Note that random cropping and transition are incorporated only for WAM, as the other methods inherently cannot survive cropping. In this section, we use \best{red} and \sbest{blue} colors to mark the best and second-best results, respectively.

\begin{table}[t]
    \caption{
        Data embedding quality compared with \textsl{cat.A} and \textsl{cat.B} methods under different levels of local tampering. Here \textsl{EdG.} represents EditGuard and the numbers in the parentheses indicate the embedding capacity in \textbf{bit}. The embedding capacity of the methods in the upper part is 324 bits.
    }\vspace{-10pt}
    \newcolumntype{M}[1]{>{\centering\arraybackslash}m{#1}}
    \renewcommand\arraystretch{0.97}
    \centering
    \small
    \scalebox{0.91}{
        \begin{tabular}{M{1.35cm}M{0.8cm}M{0.8cm}M{0.8cm}M{0.63cm}M{0.63cm}M{0.63cm}M{0.63cm}} 
        \bottomrule
        \multirow{2}{=}{\centering{\scriptsize{Method}}} & \multirow{2}{=}{\centering{\scriptsize{PSNR$\uparrow$}}} & \multirow{2}{=}{\centering{\scriptsize{SSIM$\uparrow$}}} & \multirow{2}{=}{\centering{\scriptsize{LPIPS$\downarrow$}}} &  \multicolumn{4}{c}{\scriptsize{BA~(\%) under local tampering}} \\ [-0.2pt]
        & & & & \scriptsize{15\%} & \scriptsize{30\%} & \scriptsize{45\%} & \scriptsize{60\%} \\ [-0.2pt]
        \hline

        \scriptsize{HiNet} & \scriptsize{36.345} & \scriptsize{0.7721} & \scriptsize{0.3061} & \scriptsize{\sbest{97.03}} & \scriptsize{\sbest{89.98}} & \scriptsize{\sbest{81.52}} & \scriptsize{\sbest{73.70}} \\ [-0.2pt]

        \scriptsize{ISN} & \scriptsize{38.294} & \scriptsize{\sbest{0.9426}} & \scriptsize{\sbest{0.1936}} & \scriptsize{96.30} & \scriptsize{88.40} & \scriptsize{80.72} & \scriptsize{73.04} \\ [-0.2pt]

        \scriptsize{StampOne} & \scriptsize{\sbest{40.322}} & \scriptsize{0.8836} & \scriptsize{0.2692} & \scriptsize{96.64} & \scriptsize{89.25} & \scriptsize{81.44} & \scriptsize{72.94} \\ [-0.2pt]

        \scriptsize{StegaStamp} & \scriptsize{21.217} & \scriptsize{0.7593} & \scriptsize{0.3478} & \scriptsize{92.93} & \scriptsize{85.44} & \scriptsize{77.78} & \scriptsize{69.89} \\ [-0.2pt]

        \scriptsize{Ours} & \scriptsize{\best{40.636}} & \scriptsize{\best{0.9692}} & \scriptsize{\best{0.0219}} & \scriptsize{\best{99.81}} & \scriptsize{\best{99.78}} & \scriptsize{\best{99.74}} & \scriptsize{\best{99.61}} \\ [-0.2pt]
        
        \hline

        \scriptsize{EdG.~(324)} & \scriptsize{37.238} & \scriptsize{0.8760} & \scriptsize{0.0467} & \scriptsize{99.72} & \scriptsize{99.52} & \scriptsize{97.78} & \scriptsize{97.24} \\ [-0.2pt]

        \scriptsize{Ours~(324)} & \scriptsize{40.636} & \scriptsize{0.9692} & \scriptsize{0.0219} & \scriptsize{99.81} & \scriptsize{99.78} & \scriptsize{99.74} & \scriptsize{99.61} \\ [-0.2pt]

        \scriptsize{EdG.~(64)} & \scriptsize{40.305} & \scriptsize{\sbest{0.9809}} & \scriptsize{0.0235} & \scriptsize{99.85} & \scriptsize{99.58} & \scriptsize{99.21} & \scriptsize{97.55} \\ [-0.2pt]

        \scriptsize{Ours~(64)} & \scriptsize{\sbest{40.929}} & \scriptsize{0.9717} & \scriptsize{\sbest{0.0205}} & \scriptsize{\sbest{99.86}} & \scriptsize{\sbest{99.82}} & \scriptsize{\sbest{99.81}} & \scriptsize{\sbest{99.73}} \\ [-0.2pt]

        \scriptsize{WAM~(32)} & \scriptsize{34.644} & \scriptsize{0.9223} & \scriptsize{0.0600} & \scriptsize{98.04} & \scriptsize{98.02} & \scriptsize{97.73} & \scriptsize{97.27} \\ [-0.2pt]

        \scriptsize{Ours~(32)} & \scriptsize{\best{41.132}} & \scriptsize{\best{0.9836}} & \scriptsize{\best{0.0186}} & \scriptsize{\best{99.92}} & \scriptsize{\best{99.89}} & \scriptsize{\best{99.86}} & \scriptsize{\best{99.76}} \\ [-0.2pt]

        \toprule
        \end{tabular}
    }
    \label{tab: evl_embedding_quality_cat_a}\vspace{-20pt}
\end{table}

\subsection{Data Embedding and Retrieval Quality}
\label{sec: data_embedding_quality}

Data embedding quality refers to the similarity between the host image and its corresponding stego image, whereas data retrieval quality indicates the accuracy of data decoding. We use the peak signal-to-noise ratio~(PSNR)~\cite{almohammad2010stego} which represents the pixel-level difference, SSIM~\cite{wang2004image} which reflects the structural similarity and LPIPS~\cite{zhang2018unreasonable} which indicates the perceptual quality to evaluate the stego image quality. For data decoding accuracy, we use bit accuracy~(BA, representing the percentage of correctly decoded data) for evaluation.

\subsubsection{Local Tampering Resistance}

We first compare VisGuard with \textsl{cat.A} methods under different local tampering rates~(15\% to 60\%). We use the same input data images with $18 \times 18$ data modules for these methods. The results are shown in the upper part of \cref{tab: evl_embedding_quality_cat_a}, from which we can observe the superiority of VisGuard over other methods, especially in terms of SSIM and LPIPS. This means that the stego image generated by our method has better visual similarity, which can also be supported by the qualitative comparison in \cref{fig: evl_quality_cat_a}. VisGuard can generate stego images that are almost indistinguishable from the original images. In contrast, other methods tend to introduce noticeable artifacts during this process, with some producing distortions that closely resemble the data image~(the grid-like artifacts). In practical application scenarios, this can make the existence of secret data easily noticed by users. For data decoding accuracy, when RDT and IIB are used, our method maintains a high precision under different tampering rates, whereas the accuracy of other methods are significantly affected by local tampering. Both qualitative and quantitative results reveal that, the decoding performance of \textsl{cat.A} methods nearly degraded to random guesses within tampered regions~(by observing that BA $\approx 100\%$ -- tampered rate$ \times 0.5$).

\begin{figure}[t]
    \centering
    \includegraphics[width=1.0\linewidth]{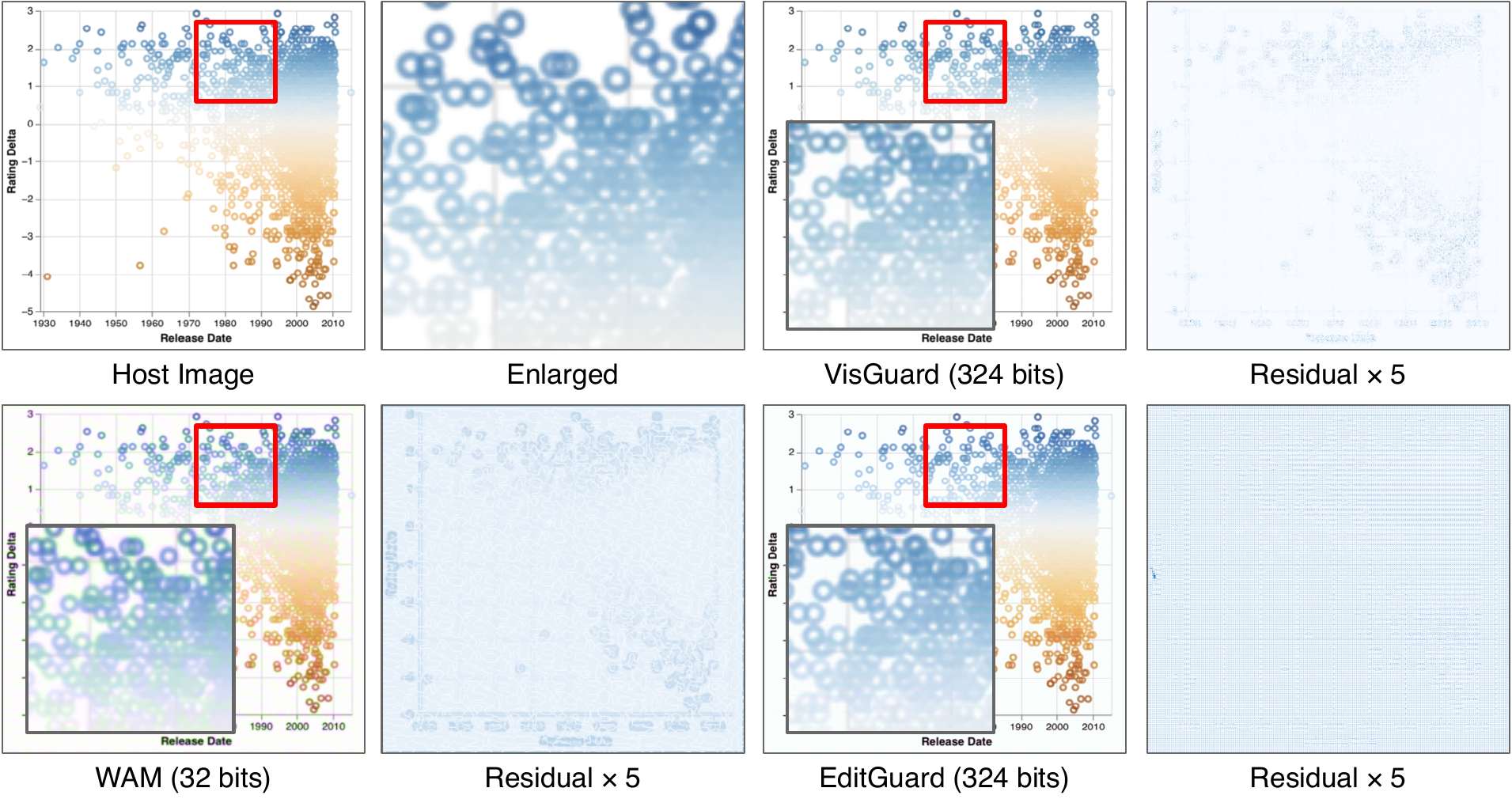}\vspace{-8pt}
    \caption{
        Stego image quality compared with \textsl{cat.B} methods. The differences are enhanced by 5 times for a better illustration.
    }
    \label{fig: evl_quality_cat_b}\vspace{-12pt}
\end{figure}

We then compare our method with \textsl{cat.B} methods. Because both WAM and EditGuard choose to embed binary data directly into host images instead of using data images as media, they have limited embedding capacity. WAM is only capable of concealing 32 bits of data, whereas EditGuard is originally designed to hide 64 bits but its model can also converge under a larger embedding capacity. Here we additionally train our model by encoding 32 bits~(with RDT(4, 8, 8, 4), which is an empirical setting) and 64 bits~(with RDT(8, 8, 4, 4)) for an intuitive comparison. A 324-bit version of EditGuard is also trained to match the capacity of our full model. The lower part of \cref{tab: evl_embedding_quality_cat_a} shows the anti-local tampering ability of \textsl{cat.B} methods. All three methods have strong resistance to local tampering~(ours is relatively better), whereas VisGuard produces significantly higher stego image quality. \cref{fig: evl_quality_cat_b} shows the stego images generated by VisGuard and \textsl{cat.B} methods, it is clear that our method produces better visual quality and that the artifacts introduced by data embedding are almost invisible. 

\subsubsection{Cropping Resistance}
\label{sec: cropping_resistance}

We exclusively compare the cropping resistance of VisGuard with that of WAM, as other methods are not designed for this type of tampering and can produce completely arbitrary decoding results~(BA $\approx$ 50\%). We first consider pure image cropping, as shown in \cref{tab: evl_cropping}, we present the BA of WAM and VisGuard under cropping rate ranging from 65\% to 95\%. Under medium cropping ratios (from 65\% to 85\%), VisGuard outperforms WAM. However, under extreme conditions, our method performs worse than WAM. This is because our method's cropping resistance is implemented by the cropping estimation introduced in \cref{sec: anchor_embedding}. In cases of severe cropping, the estimated cropping parameters can introduce errors, leading to inaccurate position rectification and resulting in low BA. However, although our full model (324-bit version) performs slightly worse, it can achieve good performance under most conditions~(cropping rate $\le 80\%$) and it has an approximately 10 times larger embedding capacity than WAM does, which makes it more suitable for practical scenarios.

\begin{table}[t]
    \caption{
        Decoding accuracy of VisGuard and WAM under cropping.
    }\vspace{-10pt}
    \newcolumntype{M}[1]{>{\centering\arraybackslash}m{#1}}
    \renewcommand\arraystretch{1.05}
    \centering
    \small
    \scalebox{0.905}{
        \begin{tabular}{M{1.9cm}M{0.63cm}M{0.63cm}M{0.63cm}M{0.63cm}M{0.63cm}M{0.63cm}M{0.63cm}} 
        \bottomrule
        \multirow{2}{=}{\centering{\scriptsize{Method}}} &  \multicolumn{6}{c}{\scriptsize{BA~(\%) under image cropping}} \\ [-0.2pt]
        & \scriptsize{65\%} & \scriptsize{70\%} & \scriptsize{75\%} & \scriptsize{80\%} & \scriptsize{85\%} & \scriptsize{90\%} & \scriptsize{95\%} \\ [-0.2pt]
        \hline

        \scriptsize{WAM~(32 bits)} & \scriptsize{97.37} & \scriptsize{96.94} & \scriptsize{96.51} & \scriptsize{95.11} & \scriptsize{\sbest{93.20}} & \scriptsize{\best{89.89}} & \scriptsize{\best{82.43}} \\ [-0.2pt]

        \scriptsize{Ours~(324 bits)} & \scriptsize{\sbest{98.79}} & \scriptsize{\sbest{98.33}} & \scriptsize{\sbest{98.27}} & \scriptsize{\sbest{95.57}} & \scriptsize{87.92} & \scriptsize{80.32} & \scriptsize{63.23} \\ [-0.2pt]

        \scriptsize{Ours~(32 bits)} & \scriptsize{\best{98.93}} & \scriptsize{\best{98.76}} & \scriptsize{\best{98.65}} & \scriptsize{\best{97.29}} & \scriptsize{\best{93.62}} & \scriptsize{\sbest{86.39}} & \scriptsize{\sbest{73.43}} \\ [-0.2pt]

        \toprule
        \end{tabular}
    }
    \label{tab: evl_cropping}\vspace{-18pt}
\end{table}

\begin{figure}[t]
    \centering
    \includegraphics[width=1.0\linewidth]{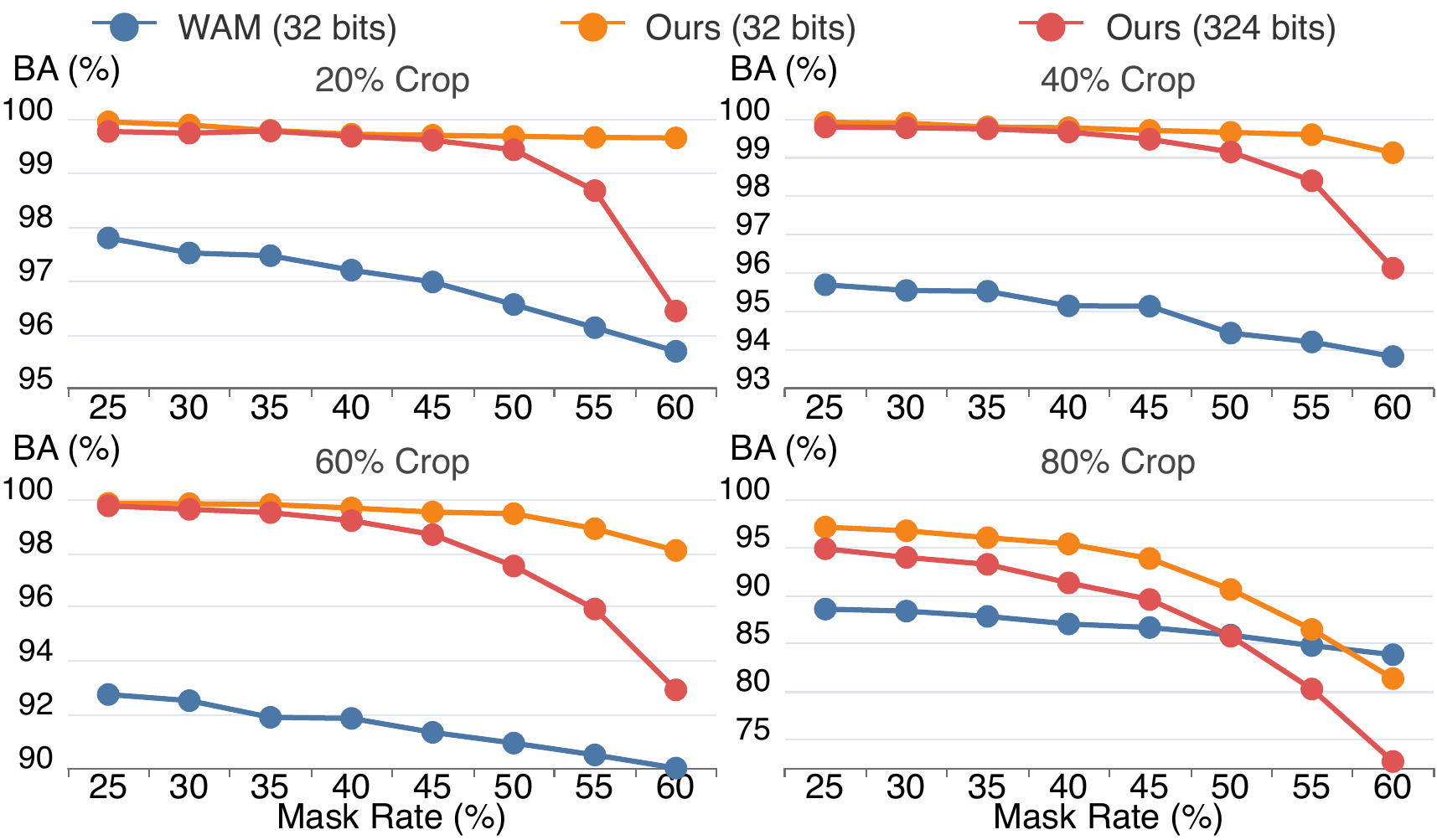}\vspace{-8pt}
    \caption{
        Bit accuracy compared with WAM under mixed tampering.
    }
    \label{fig: evl_mixed_tampering}\vspace{-12pt}
\end{figure}

Despite pure image cropping, we further evaluate the data decoding accuracy under mixed tampering, which involves performing local tampering and image cropping on stego images at the same time. We consider several levels of cropping and then apply local tampering at different rates. Note that local tampering is performed on cropped images and its ratio is relative. For example, if a stego image undergoes 60\% cropping followed by a 20\% local tampering rate, the cumulative tampering ratio is 72\% by 60\% + 60\% $\times$ 20\%. As shown in \cref{fig: evl_mixed_tampering}, our method achieves high decoding accuracy in most cases.

\begin{figure}[t]
    \centering
    \includegraphics[width=1.0\linewidth]{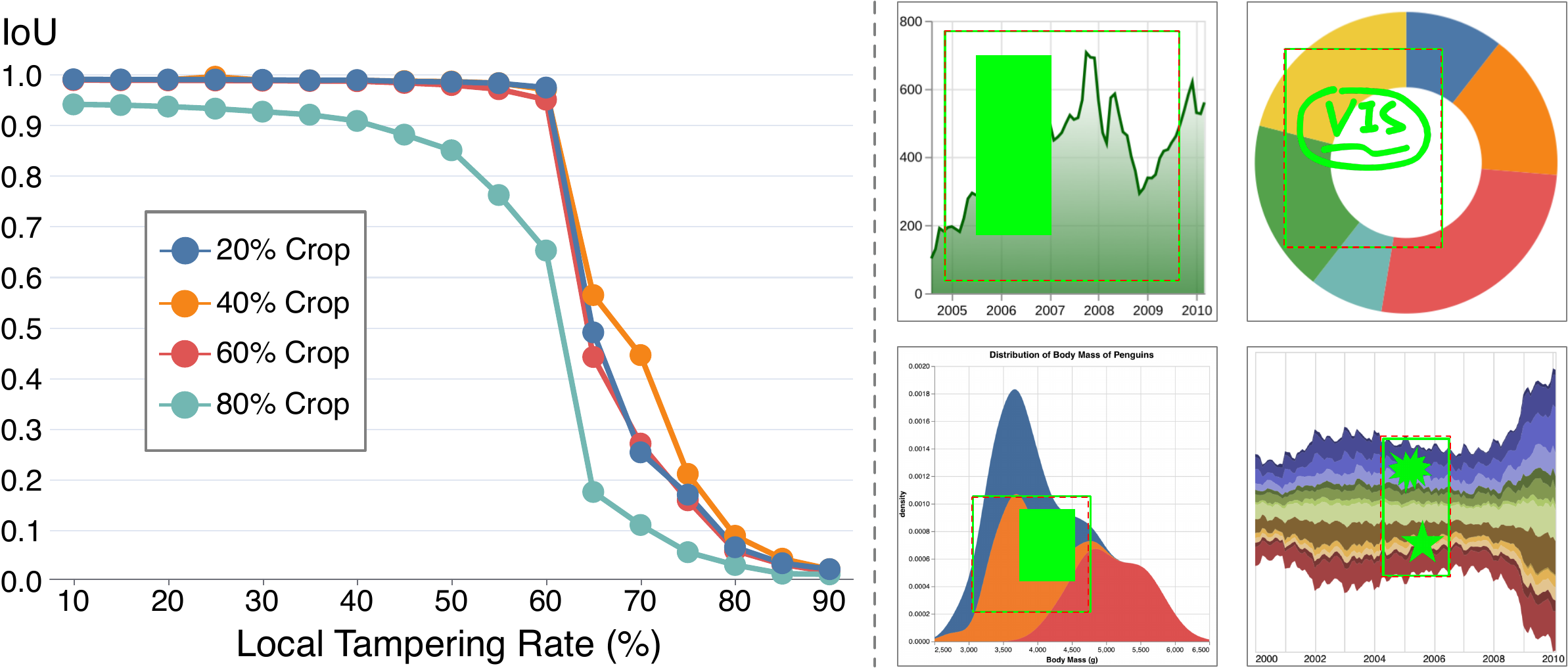}\vspace{-6pt}
    \caption{
        \textbf{Left part}: cropping estimation accuracy under mixed tampering. \textbf{Right part}: some cropping estimation results. The green frames and areas respectively denote the cropped and tampered regions under mixed tampering, while the red frames represent the estimation results.
    }
    \label{fig: evl_crop_estimation}\vspace{-20pt}
\end{figure}

We also evaluate the cropping estimation accuracy. We use IoU~(intersection over union) to measure the overlap ratio between the ground truth cropping bounding box and the predicted bounding box derived from the estimated cropping parameters. We conduct the experiment under different levels of mixed tampering and the result is shown on the left side of \cref{fig: evl_crop_estimation}. The IoU is more significantly affected by local tampering, leading to a notable decline in the IoU when the mask rate reaches 65\%. This is because local tampering impacts the local regions of the decoded anchor image, resulting in inaccurate cropping estimation. This phenomenon is particularly evident under conditions of a low cropping rate and a high masking rate. For example, when the cropping rate is 20\% (the blue line), the IoU decreases more rapidly than in the 40\% cropping scenario (the orange line). This is because local tampering occupies a larger area in the cropped image in such cases. By combining with the results in \cref{fig: evl_mixed_tampering}, it can be observed that our method has superior performance under nonextreme conditions. Some cropping estimation results are shown on the right side of \cref{fig: evl_crop_estimation}. Our method can accurately predict the cropped regions under complex mixed tampering situations.

\subsubsection{Resistance to Image Retouching and Quality Degradation}
\sidecomment{SR.4 \\ R3.2 \\ R3.3}\revision{Despite the two types of image tampering defined in \cref{sec: overview}, image retouching (e.g., brightness and contrast adjustment) and quality degradation (e.g., Gaussian noise and JPEG compression) can also interfere with the image~\cite{sahu2023study, mishra2013digital}. Although these kinds of image tampering occur less frequently in our primary focus of online image transmission scenarios, we incorporate them into our experiments to ensure a more comprehensive evaluation of VisGuard's performance.}

\revision{Given that the current experimental setup only trains the model with tampering simulation, we additionally incorporate the distortion simulation module proposed by StegaStamp~\cite{tancik2020stegastamp} to endow the model with extra robustness against image retouching and quality degradation. Specifically, we set the Gaussian noise deviation as 0.02, JPEG compression quality as 60, brightness offset as 0.1, hue shift as 0.1 and contrast range as [0.8, 1.2]. We measure the BA of different methods against such conditions and the results are shown in \cref{tab: evl_image_degradation}. VisGuard outperforms the other baselines in most cases and achieves the best overall performance. This proves that, despite the tampering types defined in \cref{sec: overview}, VisGuard also has potential resistance to other kinds of possible image tampering.}

\begin{table}[h]
    \vspace{-6pt}
    \caption{
        \revisioncaption{
            Data embedding and retrieval performance under image retouching and quality degradation. Here $\sigma$ and $\delta$ represent the Gaussian noise deviation and JPEG compression quality, respectively. \textsl{Mixed} indicates random noise combinations implemented with the distortion simulation module.}
    }\vspace{-10pt}
    \newcolumntype{M}[1]{>{\centering\arraybackslash}m{#1}}
    \renewcommand\arraystretch{1.05}
    \centering
    \small
    \scalebox{0.878}{
        \revisionbox{\begin{tabular}{M{1.4cm}M{0.7cm}M{0.7cm}M{0.7cm}M{0.63cm}M{0.63cm}M{0.63cm}M{0.63cm}M{0.75cm}} 
            \bottomrule
            \multirow{2}{=}{\centering{\scriptsize{Method}}} & \multirow{2}{=}{\centering{\scriptsize{PSNR$\uparrow$}}} & \multirow{2}{=}{\centering{\scriptsize{SSIM$\uparrow$}}} & \multirow{2}{=}{\centering{\scriptsize{LPIPS$\downarrow$}}} &  \multicolumn{5}{c}{\scriptsize{BA~(\%) under image interference}} \\ [-0.2pt]
            & & & & \scriptsize{$\sigma^{0.05}$} & \scriptsize{$\sigma^{0.1}$} & \scriptsize{$\delta^{60}$} & \scriptsize{$\delta^{20}$} & \scriptsize{Mixed} \\ [-0.2pt]
            \hline
    
            \scriptsize{~HiNet\textsuperscript{$\dagger$}} & \scriptsize{34.311} & \scriptsize{0.7666} & \scriptsize{0.4176} & \scriptsize{99.46} & \scriptsize{\best{98.70}} & \scriptsize{\sbest{99.71}} & \scriptsize{86.34} & \scriptsize{87.22} \\ [-0.2pt]
    
            \scriptsize{~ISN\textsuperscript{$\dagger$}} & \scriptsize{32.573} & \scriptsize{0.8775} & \scriptsize{0.3313} & \scriptsize{99.40} & \scriptsize{97.15} & \scriptsize{99.24} & \scriptsize{96.84} & \scriptsize{94.21} \\ [-0.2pt]
    
            \scriptsize{~StampOne\textsuperscript{$\dagger$}} & \scriptsize{\sbest{36.547}} & \scriptsize{0.8541} & \scriptsize{0.2237} & \scriptsize{99.74} & \scriptsize{98.26} & \scriptsize{99.62} & \scriptsize{\best{97.39}} & \scriptsize{\sbest{99.64}} \\ [-0.2pt]
    
            \scriptsize{~StegaStamp\textsuperscript{$\dagger$}} & \scriptsize{20.587} & \scriptsize{0.7421} & \scriptsize{0.3566} & \scriptsize{98.36} & \scriptsize{93.62} & \scriptsize{99.37} & \scriptsize{80.60} & \scriptsize{95.87} \\ [-0.2pt]
    
            \scriptsize{~EditGuard\textsuperscript{$\dagger$}} & \scriptsize{35.510} & \scriptsize{\sbest{0.9101}} & \scriptsize{\sbest{0.0692}} & \scriptsize{99.59} & \scriptsize{97.98} & \scriptsize{99.63} & \scriptsize{94.38} & \scriptsize{96.06} \\ [-0.2pt]
    
            \scriptsize{~WAM\textsuperscript{$\dagger$}} & \scriptsize{32.618} & \scriptsize{0.9018} & \scriptsize{0.0801} & \scriptsize{\sbest{99.74}} & \scriptsize{\sbest{98.41}} & \scriptsize{99.16} & \scriptsize{97.05} & \scriptsize{98.43} \\ [-0.2pt]
    
            \scriptsize{Ours\textsuperscript{$\dagger$}} & \scriptsize{\best{37.701}} & \scriptsize{\best{0.9128}} & \scriptsize{\best{0.0477}} & \scriptsize{\best{99.78}} & \scriptsize{98.38} & \scriptsize{\best{99.79}} & \scriptsize{\sbest{97.13}} & \scriptsize{\best{99.81}} \\ [-0.2pt]
    
            \toprule
            \end{tabular}}
    }
    \label{tab: evl_image_degradation}\vspace{-10pt}
\end{table}

\begin{table*}[t]
    \caption{
        \textbf{Upper part}: stego image quality and data decoding accuracy under different tampering levels of models trained with different RDT scales under 324 bits' capacity. Note that \textsl{Ours w/o RDT} is equivalent to \textsl{Ours RDT(18, 18, 1, 1)}. \textbf{Middle part}: ablation results of IIB module. \textbf{Lower part}: model performance without FEN and anchor embedding, respectively. Our full model is highlighted with \textbf{bold} font.
    }\vspace{-11pt}
    \newcolumntype{M}[1]{>{\centering\arraybackslash}m{#1}}
    \renewcommand\arraystretch{0.97}
    \centering
    \small
    \scalebox{0.91}{
        \begin{tabular}{M{4.15cm}|M{0.98cm}M{0.98cm}M{0.98cm}|M{0.75cm}M{0.75cm}M{0.75cm}M{0.75cm}M{0.75cm}|M{0.75cm}M{0.75cm}M{0.75cm}M{0.75cm}M{0.75cm}} 
        \bottomrule
        \multirow{2}{=}{\centering{\scriptsize{Method}}} & \multirow{2}{=}{\centering{\scriptsize{PSNR$\uparrow$}}} & \multirow{2}{=}{\centering{\scriptsize{SSIM$\uparrow$}}} & \multirow{2}{=}{\centering{\scriptsize{LPIPS$\downarrow$}}} & \multicolumn{5}{c|}{\scriptsize{BA~(\%) under local tampering}} & \multicolumn{5}{c}{\scriptsize{BA~(\%) under image cropping}}  \\ [-0.2pt]
        & & & & \scriptsize{30\%} & \scriptsize{45\%} & \scriptsize{60\%} & \scriptsize{75\%} & \scriptsize{90\%} & \scriptsize{50\%} & \scriptsize{60\%} & \scriptsize{70\%} & \scriptsize{80\%} & \scriptsize{90\%} \\ [-0.2pt]
        \hline

        \scriptsize{Ours w/o RDT} & \scriptsize{39.765} & \scriptsize{0.9660} & \scriptsize{0.0245} & \scriptsize{99.74} & \scriptsize{98.45} & \scriptsize{97.70} & \scriptsize{96.53} & \scriptsize{86.52} & \scriptsize{98.62} & \scriptsize{98.09} & \scriptsize{96.79} & \scriptsize{90.70} & \scriptsize{75.26} \\ [-0.2pt]

        \scriptsize{\textbf{Ours RDT(18, 18, 2, 2)}} & \scriptsize{\sbest{40.636}} & \scriptsize{\sbest{0.9692}} & \scriptsize{\sbest{0.0219}} & \scriptsize{99.78} & \scriptsize{99.74} & \scriptsize{99.61} & \scriptsize{\best{99.11}} & \scriptsize{\best{90.07}} & \scriptsize{99.12} & \scriptsize{98.80} & \scriptsize{98.33} & \scriptsize{\best{95.57}} & \scriptsize{\sbest{80.32}} \\ [-0.2pt]

        \scriptsize{Ours RDT(18, 18, 3, 3)} & \scriptsize{39.017} & \scriptsize{0.9500} & \scriptsize{0.0331} & \scriptsize{\sbest{99.83}} & \scriptsize{\sbest{99.79}} & \scriptsize{\sbest{99.66}} & \scriptsize{\sbest{98.54}} & \scriptsize{88.86} & \scriptsize{\sbest{99.31}} & \scriptsize{\sbest{99.01}} & \scriptsize{\sbest{98.39}} & \scriptsize{\sbest{95.19}} & \scriptsize{\best{81.14}} \\ [-0.2pt]

        \scriptsize{Ours RDT(18, 18, 4, 4)} & \scriptsize{37.902} & \scriptsize{0.9446} & \scriptsize{0.0432} & \scriptsize{\best{99.85}} & \scriptsize{\best{99.81}} & \scriptsize{\best{99.75}} & \scriptsize{98.30} & \scriptsize{88.12} & \scriptsize{\best{99.38}} & \scriptsize{\best{99.13}} & \scriptsize{\best{98.88}} & \scriptsize{95.13} & \scriptsize{79.00} \\ [-0.2pt]

        \hline

        \scriptsize{Ours w/o IIB} & \scriptsize{38.371} & \scriptsize{0.9532} & \scriptsize{0.0331} & \scriptsize{99.67} & \scriptsize{98.44} & \scriptsize{97.83} & \scriptsize{95.68} & \scriptsize{84.95} & \scriptsize{88.90} & \scriptsize{84.85} & \scriptsize{79.88} & \scriptsize{74.00} & \scriptsize{62.27} \\ [-0.2pt]

        \scriptsize{Ours w/o IIB RDT} & \scriptsize{36.206} & \scriptsize{0.9449} & \scriptsize{0.0482} & \scriptsize{93.22} & \scriptsize{87.76} & \scriptsize{80.18} & \scriptsize{75.13} & \scriptsize{66.25} & \scriptsize{84.68} & \scriptsize{79.13} & \scriptsize{72.91} & \scriptsize{66.31} & \scriptsize{56.92} \\ [-0.2pt]

        \hline

        \scriptsize{Ours w/o FEN} & \scriptsize{37.906} & \scriptsize{0.9512} & \scriptsize{0.0437} & \scriptsize{99.81} & \scriptsize{99.76} & \scriptsize{98.99} & \scriptsize{93.54} & \scriptsize{80.52} & \scriptsize{99.08} & \scriptsize{98.83} & \scriptsize{97.17} & \scriptsize{94.12} & \scriptsize{74.61} \\ [-0.2pt]

        \scriptsize{Ours w/o Anchor Embedding} & \scriptsize{\best{41.463}} & \scriptsize{\best{0.9751}} & \scriptsize{\best{0.0185}} & \scriptsize{99.73} & \scriptsize{99.70} & \scriptsize{98.21} & \scriptsize{97.82} & \scriptsize{\sbest{89.15}} & \scriptsize{--} & \scriptsize{--} & \scriptsize{--} & \scriptsize{--} & \scriptsize{--} \\ [-0.2pt]

        \toprule
        \end{tabular}
    }
    \label{tab: ablation}\vspace{-18pt}
\end{table*}

\begin{table}[t]
    \caption{
        Security evaluation against different steganalysis methods.
    }\vspace{-11pt}
    \newcolumntype{M}[1]{>{\centering\arraybackslash}m{#1}}
    \renewcommand\arraystretch{0.97}
    \centering
    \small
    \scalebox{0.91}{
        \begin{tabular}{M{1.5cm}|M{0.85cm}M{0.85cm}M{0.85cm}|M{0.85cm}M{0.85cm}M{0.85cm}} 
        \bottomrule
        \multirow{3}{=}{\centering{\scriptsize{Method}}} & \multicolumn{6}{c}{\scriptsize{$|$Detection Accuracy - $50\%|\downarrow$}} \\ [-0.2pt]
        & \multicolumn{3}{c|}{\scriptsize{S-UNIWARD}} & \multicolumn{3}{c}{\scriptsize{HILL}} \\ [-0.2pt]
        & \scriptsize{XuNet} & \scriptsize{KeNet} & \scriptsize{SID} & \scriptsize{XuNet} & \scriptsize{KeNet} & \scriptsize{SID} \\ [-0.2pt]

        \hline

        \scriptsize{HiNet} & \scriptsize{50.000} & \scriptsize{49.658} & \scriptsize{50.000} & \scriptsize{50.000} & \scriptsize{49.951} & \scriptsize{50.000} \\ [-0.2pt]

        \scriptsize{ISN} & \scriptsize{\sbest{9.015}} & \scriptsize{\sbest{31.583}} & \scriptsize{39.223} & \scriptsize{\best{7.836}} & \scriptsize{\sbest{27.511}} & \scriptsize{\best{10.124}} \\ [-0.2pt]

        \scriptsize{StampOne} & \scriptsize{49.962} & \scriptsize{41.729} & \scriptsize{\sbest{35.967}} & \scriptsize{42.898} & \scriptsize{43.425} & \scriptsize{\sbest{39.170}} \\ [-0.2pt]

        \scriptsize{StegaStamp} & \scriptsize{49.955} & \scriptsize{49.970} & \scriptsize{50.000} & \scriptsize{50.000} & \scriptsize{50.000} & \scriptsize{50.000} \\ [-0.2pt]

        \scriptsize{WAM} & \scriptsize{46.251} & \scriptsize{49.935} & \scriptsize{49.955} & \scriptsize{48.206} & \scriptsize{49.688} & \scriptsize{49.673} \\ [-0.2pt]

        \scriptsize{EditGuard} & \scriptsize{48.319} & \scriptsize{48.266} & \scriptsize{48.066} & \scriptsize{42.711} & \scriptsize{43.737} & \scriptsize{40.603} \\ [-0.2pt]

        \scriptsize{Ours} & \scriptsize{\best{6.561}} & \scriptsize{\best{15.516}} & \scriptsize{\best{32.818}} & \scriptsize{\sbest{27.031}} & \scriptsize{\best{8.271}} & \scriptsize{44.213} \\ [-0.2pt]
  
        \toprule
        \end{tabular}
    }
    \label{tab: evl_security}\vspace{-11pt}
\end{table}

\subsection{Security Evaluation}
\label{sec: evl_security}

Unlike stego image quality, which primarily reflects visual similarity, security mainly denotes stego images resistance to steganalysis detection. In practical scenarios, steganography methods with high security can effectively prevent malicious interception or sample leakage~\cite{yu2023cross}, ensuring that stego images are disseminated normally over the internet. In this paper, we incorporate three steganography detection methods, XuNet~\cite{xu2016structural}, KeNet~\cite{you2020siamese} and SID~\cite{tsang2018steganalyzing}, to evaluate the security of our baselines against steganalysis. Following the mainstream scheme~\cite{yu2023cross, yang2023provably, lan2023robust}, these detection models are trained on the BOSSbase 1.01 dataset~\cite{bas2011break} with S-UNIWARD~\cite{holub2013digital} and HILL~\cite{li2014new} embedding, respectively.

During evaluation, for one host image, we jointly feed itself and its stego image embedded with random data to the detection model. For steganography methods with higher security, the detection accuracy should be closer to 50\%, meaning that the detection model cannot distinguish between host images and stego images, causing the results to approximate random guessing. The results in \cref{tab: evl_security} demonstrate that VisGuard has higher security than the other methods do in most cases. Although ISN sometimes performs better, as shown in \cref{fig: evl_quality_cat_a}, it can cause obvious artifacts in stego images and the existence of secret data can be easily detected by human eyes.

\begin{table}[t]
    \caption{
        Performance of \textsl{cat.A} methods trained with RDT(18, 18, 2, 2).
    }\vspace{-11pt}
    \newcolumntype{M}[1]{>{\centering\arraybackslash}m{#1}}
    \renewcommand\arraystretch{0.95}
    \centering
    \small
    \scalebox{0.91}{
        \begin{tabular}{M{1.35cm}M{0.8cm}M{0.8cm}M{0.8cm}M{0.63cm}M{0.63cm}M{0.63cm}M{0.63cm}} 
        \bottomrule
        \multirow{2}{=}{\centering{\scriptsize{Method}}} & \multirow{2}{=}{\centering{\scriptsize{PSNR$\uparrow$}}} & \multirow{2}{=}{\centering{\scriptsize{SSIM$\uparrow$}}} & \multirow{2}{=}{\centering{\scriptsize{LPIPS$\downarrow$}}} &  \multicolumn{4}{c}{\scriptsize{BA~(\%) under local tampering}} \\ [-0.2pt]
        & & & & \scriptsize{15\%} & \scriptsize{30\%} & \scriptsize{45\%} & \scriptsize{60\%} \\ [-0.2pt]
        \hline

        \scriptsize{~HiNet\textsuperscript{$\ddagger$}} & \scriptsize{38.286} & \scriptsize{0.8277} & \scriptsize{0.2673} & \scriptsize{\sbest{99.21}} & \scriptsize{\sbest{98.98}} & \scriptsize{\sbest{{98.76}}} & \scriptsize{\sbest{96.73}} \\ [-0.2pt]

        \scriptsize{~ISN\textsuperscript{$\ddagger$}} & \scriptsize{40.084} & \scriptsize{0.9583} & \scriptsize{0.1514} & \scriptsize{97.99} & \scriptsize{97.77} & \scriptsize{96.44} & \scriptsize{95.95} \\ [-0.2pt]

        \scriptsize{~StampOne\textsuperscript{$\ddagger$}} & \scriptsize{\best{41.576}} & \scriptsize{\sbest{0.9650}} & \scriptsize{\sbest{0.1409}} & \scriptsize{99.17} & \scriptsize{\sbest{98.98}} & \scriptsize{98.74} & \scriptsize{96.56} \\ [-0.2pt]

        \scriptsize{StegaStamp\textsuperscript{$\ddagger$}} & \scriptsize{23.223} & \scriptsize{0.7900} & \scriptsize{0.3250} & \scriptsize{97.27} & \scriptsize{96.03} & \scriptsize{95.11} & \scriptsize{93.96} \\ [-0.2pt]

        \scriptsize{Ours} & \scriptsize{\sbest{40.636}} & \scriptsize{\best{0.9692}} & \scriptsize{\best{0.0219}} & \scriptsize{\best{99.81}} & \scriptsize{\best{99.78}} & \scriptsize{\best{99.74}} & \scriptsize{\best{99.61}} \\ [-0.2pt]

        \toprule
        \end{tabular}
    }
    \label{tab: evl_cat_a_rdt}\vspace{-20pt}
\end{table}

\subsection{Ablation Study}

\textbf{Repetitive Data Tiling} To validate the effectiveness of RDT, we train our model with different RDT scales under the same embedding capacity of 324 bits. As shown in the upper part of \cref{tab: ablation}, compared with \textsl{Ours w/o RDT}, RDT can enhance both stego image quality and decoding accuracy. However, as the RDT scale increases, the metrics begin to decrease, with only partial improvements in some cases. This is because increasing the RDT scale leads to denser data blocks in the tiled data image, which places a greater burden on the network's ability to learn feature representations, thereby resulting in an overall performance drop. Our full model, which employs RDT(18, 18, 2, 2), achieves relatively balanced performance. In addition, because RDT can be easily applied to \textsl{cat.A} methods that use images to represent secret data, we further train these methods with the same RDT preprocessing as VisGuard's. The results are shown in \cref{tab: evl_cat_a_rdt}. All these methods show performance improvements over the results in \cref{tab: evl_embedding_quality_cat_a}. This proves the effectiveness and scalability of RDT for tamper-resistant data embedding.

\textbf{Invertible Information Broadcasting} We evaluate the effectiveness of the IIB module by training our model without using it. The results are shown in the middle part of \cref{tab: ablation}. The IIB module can not only improve the stego image quality but also significantly increase the decoding accuracy, especially when facing image cropping. We additionally train a model without using both IIB and RDT~(last row of \cref{tab: ablation}), and the results demonstrate that these two modules can collectively enhance the overall performance of VisGuard.

\textbf{Feature Enhancement Network} We remove the FEN and then retrain our model. The results shown in the lower part of \cref{tab: ablation} reveal that the FEN can effectively improve the data embedding quality. This is because the FEN can enhance features weakened after tampering, thereby alleviating the learning burden on the backbone network and improveing its performance.

\sidecomment{R1.7}\revision{\textbf{Anchor Embedding} We also test the model performance without the anchor embedding network. The results are shown in the lower part of \cref{tab: ablation}. Although anchor embedding makes VisGuard cropping-resistant, it slightly reduces the stego image quality.}

\begin{figure}[t]
    \centering
    \revisionbox{\includegraphics[width=1.0\linewidth]{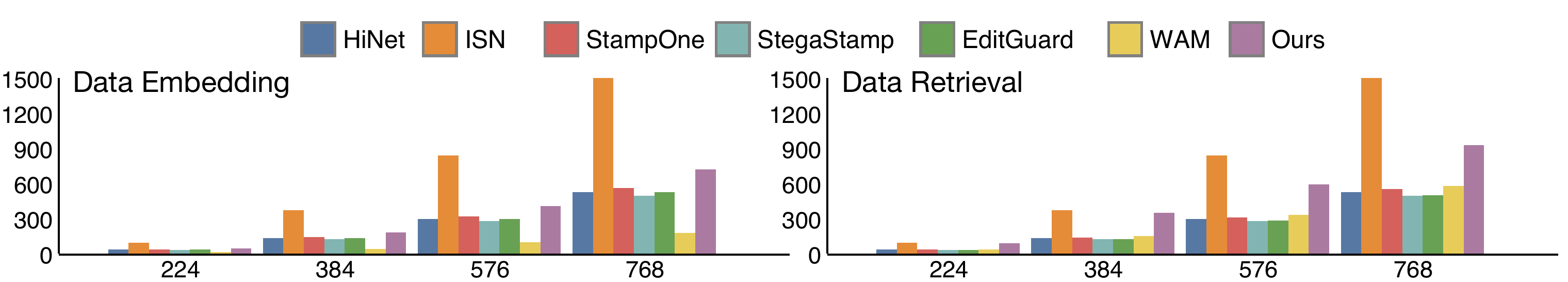}}\vspace{-8pt}
    \caption{
        \revisioncaption{Calculation cost at different image resolutions measured by \textbf{GFlops}.}
    }
    \label{fig: evl_performance}\vspace{-8pt}
\end{figure}

\subsection{Limitation Analysis}
\label{sec: limitation}

\sidecomment{SR.5 \\ R2.2 \\ R3.5}The current version of VisGuard still has some limitations. \revision{First, using anchor embedding for cropping resistance can introduce extra computations. As shown in \cref{fig: evl_performance}, our method's calculation cost is subject to image resolution. Although using the additive stego watermark introduced in \cref{sec: additive_stego_watermark} can reduce the computational cost for high-resolution images in practical scenarios, VisGuard's inherent computational efficiency remains suboptimal.}

\begin{figure}[t]
    \centering
    \includegraphics[width=1.0\linewidth]{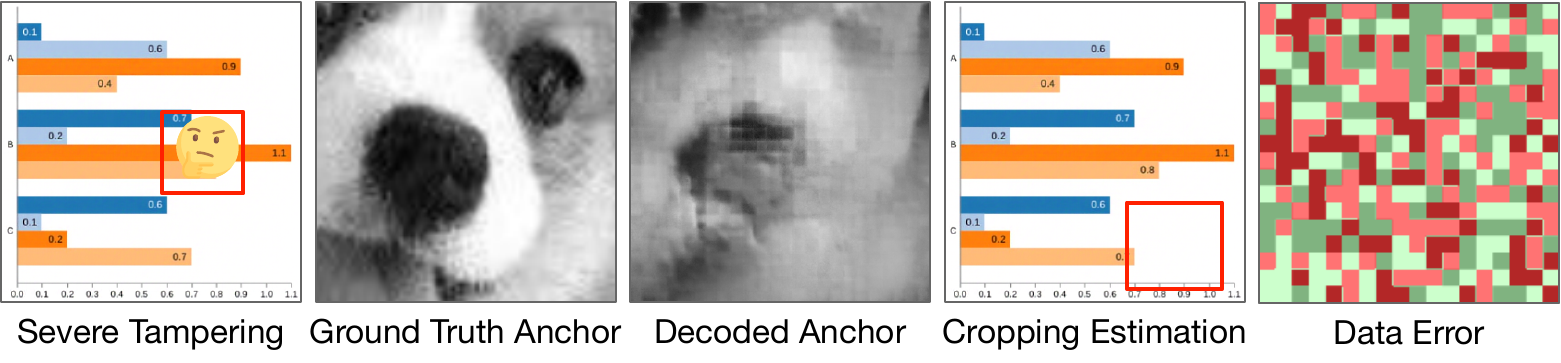}\vspace{-10pt}
    \caption{
        Failure case: severe tampering~(in this example, 95\% cropping + 60\% local tampering) can cause incorrect cropping estimation.
    }
    \label{fig: evl_limitation}\vspace{-18pt}
\end{figure}

However, VisGuard's ability to withstand extreme tampering is still insufficient, which is primarily due to the bottleneck caused by cropping estimation, as discussed in \cref{sec: cropping_resistance}. As shown in \cref{fig: evl_limitation}, under such conditions, the decoded anchor image is largely affected, which leads to incorrect cropping parameter prediction and a low BA. In contrast, WAM is more competent in such cases, but its embedding capacity is limited to 32 bits, which is not sufficient for practical application.

Another shortcoming of VisGuard is its insufficient embedding capacity compared with previous steganography-based VIDR methods, i.e., VisCode~\cite{zhang2020viscode} and InvVis~\cite{ye2023invvis}, a comparison is shown in \cref{tab: evl_capacity}. These two methods choose to embed all the data required within images, which is useful under offline decoding situations, whereas we choose to embed a link instead to achieve robustness against tampering.

\begin{table}[h]\vspace{-8pt}
    \caption{
        Embedding capacity measured by BPP~(bit per pixel)~\cite{ye2023invvis} of different steganography-based methods. Here \textsl{EdG.} represents EditGuard.
    }\vspace{-8pt}
    \newcolumntype{M}[1]{>{\centering\arraybackslash}m{#1}}
    \renewcommand\arraystretch{1.05}
    \centering
    \small
    \scalebox{1.0}{
        \begin{tabular}{M{1.5cm}|M{0.95cm}M{0.95cm}M{0.95cm}M{0.95cm}M{0.95cm}} 
        \bottomrule
         \scriptsize{Method} & \scriptsize{VisCode} & \scriptsize{InvVis} & \scriptsize{WAM} & \scriptsize{~EdG.} & \scriptsize{Ours} \\ [-0.2pt]
        \hline
        \scriptsize{~BPP$\times$100 $\uparrow$} & \scriptsize{\sbest{0.643}} & \scriptsize{\best{1.295}} & \scriptsize{0.016} & \scriptsize{0.008} & \scriptsize{0.073} \\ [-0.2pt]
        \toprule
        \end{tabular}
    }
    \label{tab: evl_capacity}\vspace{-18pt}
  \end{table}

\section{Future Work and Conclusion}

Although promising, VisGuard still has room for improvement. \sidecomment{SR.4 \\ R1.2 \\ R3.3}\revision{For example, as suggested by Fu et al.~\cite{fu2022chartstamp}, further enhancing the robustness of our method against severe distortions such as color quantization and perspective transformation can improve its practical applicability.} In addition, improving VisGuard's resistance to extreme tampering and computational efficiency are also effective ways to improve practicability.

In summary, VisGuard is a tamper-resistant VIDR framework that supports various application scenarios, such as robust invertible visualization, visualization tampering detection and localization and source-end visualization data embedding. We propose a deep steganography-based pipeline for high-quality data embedding. We propose repetitive data tiling and invertible information broadcasting to increase the robustness of our method. We also outline a new cropping resistance and localization scheme that leverages the encoding of extra anchor images. Experiments demonstrate that VisGuard can achieve high-quality, high-security, and high-capacity data embedding compared with previous methods. To our best knowledge, VisGuard is the first effort to address the tamper-resistant issue in the context of visualization image data retrieval. We believe that our approach holds strong potential for practical applications. 


\acknowledgments{%
The authors wish to acknowledge the support from the Natural Science Foundation of Shanghai Municipality, China under Grant 24ZR1418300.
}

\bibliographystyle{abbrv-doi-hyperref}

\bibliography{template}







\end{document}